\pdfoutput=1
\documentclass[11pt]{article}

\usepackage[]{acl}

\usepackage{times}
\usepackage{latexsym}

\usepackage{subcaption}
\usepackage{graphicx}
\usepackage{booktabs}
\usepackage{multirow}
\usepackage{color}
\usepackage{colortbl}
\usepackage{listings}
\usepackage{amsmath}
\usepackage{soul}
\usepackage{svg}
\usepackage{pifont}

\usepackage{pgfplots}

\newcommand{\Firstaspect}{Retrieval robustness in case of table structure changes}
\newcommand{\Secaspect}{Attention to relevant cells} %Robustness  for attention to table content
\newcommand{\Thirdaspect}{Aggregation/comparison robustness in case of value changes}
\usepackage[T1]{fontenc}
\usepackage[utf8]{inputenc}

\usepackage{microtype}

\usepackage{inconsolata}

\usepackage{xspace}
\usepackage{graphicx}
\usepackage{enumitem}
\usepackage{xcolor}
\usepackage{tikz-dependency}
\usetikzlibrary{shapes,arrows}
\usepackage{amssymb}
\usepackage{float}
\usepackage{todonotes}

\usepackage{hyperref}       % hyperlinks
\usepackage{url}            % simple URL typesetting
\usepackage{booktabs}       % professional-quality tables
\usepackage{amsfonts}       % blackboard math symbols
\usepackage{nicefrac}       % compact symbols for 1/2, etc.
\usepackage{natbib}
\usepackage{doi}

\usepackage{caption}
\usepackage{subcaption}

\usepackage{lscape}
\usepackage{longtable} % tables on multiple pages
\usepackage{multirow}
\usepackage{dsfont}

\setlist[description]{leftmargin=\parindent,labelindent=0pt}

\definecolor{darkgreen}{rgb}{0, 0.75, 0}
\newcommand{\red}[1]{\textcolor{red}{#1}}

\newcommand{\enquote}[1]{``#1''}

\newcommand{\aref}[1]{Appendix~\ref{#1}}

\newcommand{\tref}[1]{Table~\ref{#1}}
\newcommand{\fref}[1]{Figure~\ref{#1}}

\definecolor{talita}{rgb}{0.635,0.998,0.722}
\definecolor{anne}{rgb}{0.8,0.8,1}
\definecolor{wei}{rgb}{0.998,0.722,0.635}
\definecolor{heike}{rgb}{0.4, 0.8, 0.4}
\definecolor{final}{rgb}{1, 1, 0.6}

\usepackage{color}
\usepackage{soul}

\newcommand{\MM}[2][]{\textcolor{black}{#2}}

\newcommand{\wei}[2][]{\textcolor{black}{#2}}

\newcommand{\heike}[2][]{\textcolor{black}{#2}}

\newcounter{example}

\usepackage{array}
\newcolumntype{L}[1]{>{\raggedright\let\newline\\\arraybackslash\hspace{0pt}}m{#1}}
\newcolumntype{C}[1]{>{\centering\let\newline\\\arraybackslash\hspace{0pt}}m{#1}}
\newcolumntype{R}[1]{>{\raggedleft\let\newline\\\arraybackslash\hspace{0pt}}m{#1}}
\usepackage{rotating}

\title{FREB-TQA: A Fine-Grained Robustness Evaluation Benchmark\\ for Table Question Answering}

\author{Wei Zhou$^{1,3}$ \hspace{5mm}
  Mohsen Mesgar$^1$ \hspace{5mm}
  Heike Adel$^{2}$ \hspace{5mm}
  Annemarie Friedrich$^3$ \\
  $^1$Bosch Center for Artificial Intelligence, Renningen, Germany \\ 
    $^2$Hochschule der Medien, Stuttgart, Germany \hspace{5mm} $^3$University of Augsburg, Germany \hspace{2.0mm} \\
\texttt{\{wei.zhou|mohsen.mesgar\}@de.bosch.com}\\ %\hspace{5mm} \texttt{adel-vu@hdm-stuttgart.de}\\
  \texttt{annemarie.friedrich@informatik.uni-augsburg.de}}

\begin{document}
\maketitle
\begin{abstract} %\heiketodo{I added brackets to Wei's and Mohsen's email addresses to make it look like a regex -- this is how this is typically formatted}
Table Question Answering (TQA) aims at composing an answer to a question based on tabular data. 
\MM[While previous work has shown that TQA models suffer from a lack of robustness,  the cause and nature of these issues are still largely unclear, which is a key bottleneck for developing robust TQA systems. ]
{While prior research has shown that TQA models lack robustness, understanding the underlying cause and nature of this issue remains predominantly unclear, posing a significant obstacle to the development of robust TQA systems.}
\heike[
%\MM[
In this paper, we introduce three aspects of robustness and create a novel benchmark in English for the fine-grained robustness evaluation of TQA systems: (1) System responses should be invariant to changes in table structure; (2) systems should rely on the content of the relevant cells when composing their answers and avoid answering using their biases; and (3) they should have robust numerical reasoning capability.  %]
%{
In this paper, we formalize three major aspects of robustness in TQA systems. 
They should (i) answer questions regardless of alterations in table structure, (ii) base their responses on the content of relevant cells rather than on biases, and (iii) demonstrate robust numerical reasoning capabilities. 
Accordingly, we establish a novel benchmark in English for the fine-grained evaluation of robustness in TQA systems.
%}
]{In this paper, we formalize three major desiderata for a fine-grained evaluation of robustness of TQA systems. They should (i) answer questions regardless of alterations in table structure, (ii) base their responses on the content of relevant cells rather than on biases, and (iii) demonstrate robust numerical reasoning capabilities. To investigate these aspects, we create and publish a novel TQA evaluation benchmark in English.}
\MM[Our extensive experimental analysis of state-of-the-art TQA systems shows that no system consistently excels in these three aspects.]
{
Our extensive experimental analysis reveals that none of the examined state-of-the-art TQA systems consistently excels in these three aspects. 
}
Our benchmark is a crucial instrument for monitoring the behavior of TQA systems and paves the way for the development of robust TQA systems.
We release our benchmark publicly.\footnote{\url{https://github.com/boschresearch/FREB-TQA}} %\wei[]{at \href{https://github.com/boschresearch/FREB-TQA}{https://github.com/boschresearch/FREB-TQA}.}
%\heiketodo{please create a git repo at the Bosch research github and link it here!}
%\weitodo{can i firstly create a git repo without any content to fill this place holder? Then the content can be added once the open source is finished.}}.\heiketodo{yes, you can. pls do so!\MMN{Agree!}}

\end{abstract}

   \section{Introduction}
Table Question Answering (TQA) deals with answering natural language questions related to information organized in a table.  
TQA systems serve as a fundamental component for interacting with relational databases \cite{Zhong2017Seq2SQLGS,yu-etal-2018-spider} through natural language and for processing information across diverse domains, e.g., science \cite{Desai2021TabLeXAB} and finance \cite{Zhu2021TATQAAQ}. 

Processing tabular knowledge poses notable challenges. 
While tables are structured, there are no standard table layouts for representing a particular type of data.
\MM[Table cells may include different data types such as text and numbers, and table structure may be nested.]
{Table cells may contain various data types, including text and numbers. 
Moreover, tables can have nested structures.}
\MM[This requires TQA systems to combine text-based commonsense with numerical reasoning over structured data.]
{Thus, TQA systems should be able to robustly integrate textual commonsense understanding with numerical reasoning applied to structured data.}

% \annetodo{TODOs for teaser image: remove abbreviations for aspect, spell out where necessary, Floor $\rightarrow$ Floors, you can save some space by puttin ghte two bottom tables next to each other. the red arrows may be a bit confusing, it looks like the bottom tables are somehow then going up, but that doesn't happen in the benchmark}

\begin{figure}[!t]
    \centering
    \includegraphics[width=1.\columnwidth]{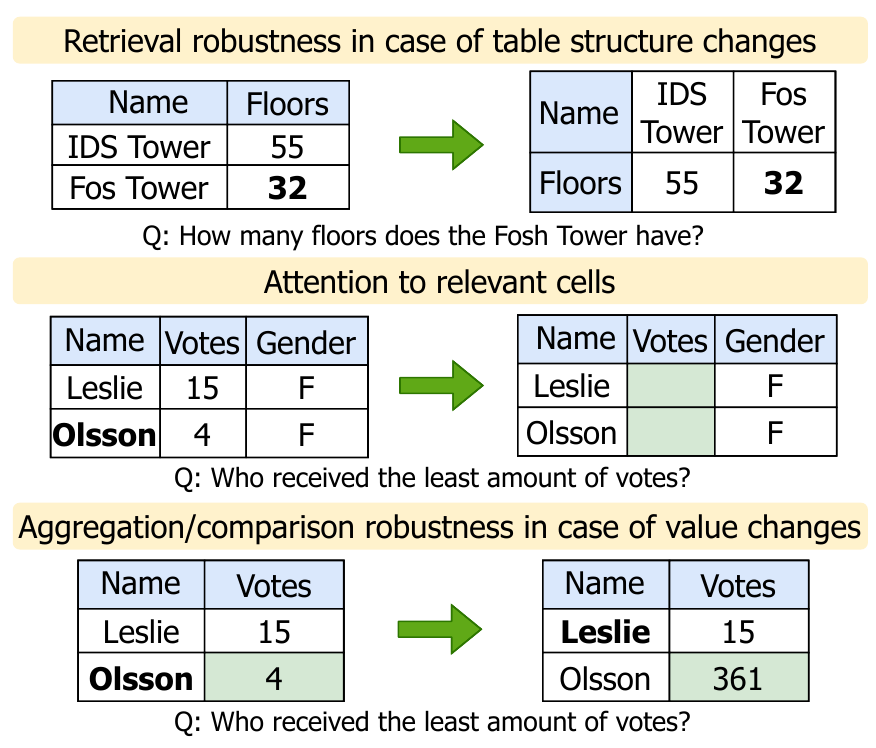}
    \caption{
    Our benchmark addresses three aspects of robustness shown in the yellow boxes. 
    Answers are bold in tables.
    Original tables (left) are what exist in a TQA dataset and changed tables (right) show tables after perturbations. We demonstrate three perturbations in this figure (top to bottom): table transposing, removing relevant cells and modifying values to change answers.}
    \label{fig:framework}
\end{figure}

Recent benchmarks for TQA systems \cite{Zhao2023RobuTAS,Yang2022TableFormerRT} reveal that current TQA systems generate inconsistent responses by performing coarse-grained changes in tables and questions.
However, these benchmarks contain a mix of questions requiring different levels of retrieval and aggregation \wei[complexities]{capabilities}.
%\MMN{What is comparing outputs? outputs vs inputs?}
 
%\heiketodo{you require an ``ability''/``capability'' not a ``complexity''} \MMN{I agree! it should be capability!} 
\MM[As a result, they do not distinguish between different aspects of (non-)robustness, e.g., which errors are due to a failure in retrieving the correct cell or which are due to a failure in the value aggregation steps.]
{As a result, these benchmarks fail to differentiate between various aspects of robustness, such as errors stemming from incorrect cell retrieval versus those arising from failures in value aggregation steps.}
Pinpointing the exact failures of \MM[]{TQA} systems, however, is a necessary diagnostic step towards enhancing their robustness. 

In this paper, we lay the groundwork for a more fine-grained systematic evaluation of TQA systems. 
%\weitodo{"as illustrated in Figure 1", this phrase does not go well with the rest of the paragraph as Figure 1 only provides some straightfoward presentation of the perturbation}
% \blue{There are many aspects to robustly answering questions based on table data.
\heike[We]{To foster the development of robust TQA systems, we} address three\footnote{We acknowledge that there are additional aspects that future work should address.} important desiderata\heike[for robust TQA systems]{}, as illustrated in \fref{fig:framework}.
%\annetodo{footnote: I would either mention several examples or none}
%\MM[Our set of aspects intends]{These aspects intend} 
\heike[]{With these aspects, we intend} to separate several steps required when answering questions based on tabular data, i.e., identifying the relevant table cells and aggregating over or comparing them.
%\wei[]{These aspects are inspired by how questions related to tables are solved [Add citation for both human and systems here]\cite{}, i.e.,  retrieving relevant information and aggregating over/comparing them.} 
\heike[To foster the development of robust TQA systems]{In particular}, we propose to evaluate \heike[them]{TQA systems} according to three aspects:
(1) \textbf{\Firstaspect}: Systems should retrieve the correct relevant cells regardless of table structure changes;
(2) \textbf{\Secaspect}: The system should use the relevant cell values from retrieval instead of exploiting shortcuts, such as model-internal knowledge or positional biases, when composing answers; and 
(3) \textbf{\Thirdaspect}: The system should aggregate the values of relevant cells correctly regardless of cell value changes. 
Aggregation can involve different types of reasoning. In this study, 
we focus on numerical operations, such as counting and comparisons, as they are the most common reasoning types in TQA \cite{Zhu2021TATQAAQ}.
%reasoning, \wei[]{reflected by operations such as} \st{including}
%counting, comparing, \wei[]{as it is one of the most common reasoning in TQA \cite{Zhu2021TATQAAQ}.}

To evaluate TQA systems with regard to these aspects, we create \textbf{FREB-TQA}, a new benchmark based on four well-studied TQA datasets using seven novel automatic perturbations, one perturbation from previous work \cite{Zhao2023RobuTAS}, and extensive manual annotations. 
\MM[The methodology used for constructing our benchmark demonstrates how to obtain high-quality data for evaluating TQA systems in a fine-grained way.]{}

In our extensive experiments, we study pipeline and end-to-end state-of-the-art TQA systems \cite{Liu2021TAPEXTP,jiang-etal-2022-omnitab,Herzig2020TaPasWS,Cheng2022BindingLM}, as well as large language models (LLMs) \cite{Touvron2023Llama2O} using our \MM[new]{} benchmark.
\MM[The fine-grained results reveal shortcomings of these systems that have not been explored before.]
{The fine-grained results uncover shortcomings of these systems that have not been previously explored.}
For example, we show that model performance diminishes substantially when column order changes, or when cells containing the answer are positioned at the bottom of the table.
\MM[This finding indicates a strong positional bias in TQA systems, motivating the need for more informative table encodings.]{
This finding indicates a strong positional bias in TQA systems, motivating more informative table encodings.
}
%We further show that at most 9\% of questions in the previous coarse-grained benchmarks can be answered by TQA systems without accessing tables. 
%We argue that such questions are not adequate to benchmark TQA systems.
% In our paper, fine-grained generative TQA systems exhibit the best performance and least drops when applied in-domain, but on TAT, which includes numerical reasoning, a hybrid system \cite{Cheng2022BindingLM} that may include calls to Python or databases, performs much better, yet still with an accuracy of only around 65\% on recent benchmarks.
% \annetodo{correct? the difference on this dataset would be good to highlight, not sure if that is still possible. I am referring to Table 4 here} \weitodo{I think it is good to highlight the difference between in-domain or not, but I am confused where does the number 65\% come from?}\annetodo{I took that from their paper, please double-check}
% We believe that using computational tools for structured data is the most promising direction, as our study suggests that it may be hard to achieve robustness for the TQA task with purely generative models.
% Our benchmark provides an instrument for analyzing, and hence improving, TQA systems to make them more robust.

Our main contributions are: 
\textbf{(1)} We propose a novel benchmark of 8,590 selected and partially manually annotated questions and tables, resulting in a total of 75,205 instances using seven perturbation methods;  
\textbf{(2)} using several inter-annotator agreement studies, we demonstrate the solidness of our benchmark; and
\textbf{(3)} using our benchmark, we experiment extensively with state-of-the-art TQA systems, discovering that for almost all robustness tests, system performances drop, demonstrating the difficulty of our benchmark for TQA systems.
    \section{Related Work}
\label{sec:rel}

\MM[In this section, we provide an overview of related work on TQA systems and on evaluating their robustness.]
{We provide an overview of previous research on TQA systems and their evaluation for robustness.}

\paragraph{TQA systems.}%\annetodo{to make this paper readable in the future: maybe add one sentence that says that tables are serialized into some string that gets fed into the models in a prompting style - I feel this is kind of crucial, and who knows that type of systems will be well-known in a couple of years} %There are common systems.
End-to-end TQA systems \cite{Liu2021TAPEXTP,Jiang2022OmniTabPW} compose an answer given a question and a serialized version of the table without intermediate steps or using additional tools. %\annetodo{if space: extend explanation, mention pre-training and fine-tuning stages (1-2 sentences!)}
% State-of-the-art End-to-end systems systems include TAPEX \cite{Liu2021TAPEXTP} and OmniTab \cite{Jiang2022OmniTabPW}. 
% They have not been trained on unanswerable questions.
Pipeline systems convert questions into a command language, e.g., SQL \cite{Cheng2022BindingLM, Ni2023LEVERLT}, filtered tables \citep{Glass2021CapturingRA, lei-etal-2022-assistsr,Ye2023LargeLM,Herzig2020TaPasWS}, and then generate answers either via executing the commands \citep{Ni2023LEVERLT,Zhang2023ReAcTableER} or via using a trained neural network \citep{Lei2023S3HQAAT}. 
Our benchmark reveals the key strengths and weaknesses of these types of systems.

\paragraph{Robustness evaluation.}
Several studies have shown that recently proposed TQA systems suffer from robustness issues \cite{Yang2022TableFormerRT,Zhao2023RobuTAS,Lin2023AnIT}.
\citet{Zhao2023RobuTAS} provide a robustness evaluation dataset for TQA, which includes header perturbations, content perturbations, and question perturbations.
Their work is closely related to ours. However, they do not further disentangle the various aspects of robustness, thus not providing detailed insights into why systems are not robust.
\MM[The finer-grained aspects targeted by FREB-TQA can help to gain a deeper understanding of TQA systems and to identify possible improvements of TQA systems in a targeted manner.] %\heiketodo{``in a systematic way'' (instead of ``targeted manner'') as the sentence contains ``targeted'' twice!}
{As we show in our experiments, the fine-grained aspects formulated in FREB-TQA contribute to a deeper understanding of TQA systems.}

%\green{We are also aware of several studies \wei[addressing table-related natural language processing tasks other than TQA.]{looking into the proposed aspects in tasks other than TQA.}
\wei[]{To the best of our knowledge, no previous work on TQA provides analyses of models for our proposed aspects. However, there are several studies looking into those aspects for other tasks.}
%\MMN{Instead of "We are also aware of several studies", say why the following papers should be discussed. }
In the context of tabular natural language inference, \citet{gupta-etal-2022-model,Gupta2022RightFT} study to what extent models pay attention to relevant cells.
%aggregation and comparison robustness in case of value changes.
The robustness of large language models \heike[to perform calculations correctly]{} 
in case of numerical value changes has primarily been studied in the context of solving math world problems \cite{Stolfo2022ACF} and tabular natural language inference \cite{Akhtar2023ExploringTN}.

%When zooming into specific aspects of robustness, i.e., attention to relevant cells, \wei[]{focusing on tabular NLI} \citep{gupta-etal-2022-model,Gupta2022RightFT}, 
%aggregation/comparison robustness in case of value changes, \wei[]{focusing on either tabular NLI or math word problems} \weitodo{reads a bit weird}
%\cite{Akhtar2023ExploringTN,Stolfo2022ACF}, \red{none of them are studied in TQA.}
%\annetodo{I do not understand this: the papers you cite here are doing precisely this?}

%\annetodo{need to cite!! https://arxiv.org/pdf/2311.02216.pdf}

%\citet{Stolfo2022ACF} evaluate numerical reasoning abilities of models on mathematical problems. 
%However, numerical reasoning in TQA requires systems to understand table structure and can process long contexts.  
%\MMN{How does your work differ from the above papers?
%shortcut usages: Right for the Right Reason: Evidence Extraction for Trustworthy Tabular Reasoning, 
%Is My Model Using the Right Evidence? Systematic Probes for Examining
%Evidence-Based Tabular Reasoning
%Lost in the Middle: How Language Models Use Long Contexts

 \section{Our Benchmark: \textsc{FREB-TQA}}
\label{sec:benchmark}

\begin{table}[!t]
    \small
    \centering
    \begin{tabular}{@{}lccc@{}}
         \toprule
         \textbf{Source Dataset}& \textbf{\# EQs} & \textbf{\# RQs} &
         \textbf{\# ORI}\\
         \midrule
         WTQ \cite{Pasupat2015CompositionalSP} & 205 & 1562 & 2831\\
         WikiSQL \cite{Zhong2017Seq2SQLGS} & 6013 & 0 & 8418 \\ 
         SQA \cite{Iyyer2017SearchbasedNS}  & 157 & 0 &2265 \\
         TAT \cite{Zhu2021TATQAAQ} & 114 & 539 &1668\\ 
         \bottomrule
    \end{tabular}
    \caption{The first two columns show the total number of questions selected from \heike[]{the dev part of} each source dataset for extraction questions (EQs) and reasoning questions (RQs). \wei[]{The last column shows the original number of questions in each source dev set.}}
    \label{tab:source_dataset_contr}
\end{table}

\MM[We construct FREB-TQA, our new \textbf{F}ine-grained \textbf{R}obustness \textbf{E}valuation \textbf{B}enchmark for \textbf{T}able \textbf{Q}uestion  \textbf{A}nswering.]{
FREB-TQA is a \textbf{F}ine-grained \textbf{R}obustness \textbf{E}valuation \textbf{B}enchmark for \textbf{T}able \textbf{Q}uestion  \textbf{A}nswering.}
From \MM[a set of source] {four TQA} datasets (\MM[]{Table \ref{tab:source_dataset_contr}}), we first classify questions with regard to whether questions merely require cell value retrieval or further reasoning.
\MM[Table \ref{tab:source_dataset_contr} shows the final number of questions each source dataset contributes to our benchmark. ]{}
%\heiketodo{In the table caption: change ``dev set'' to ``dataset''? or: change to: ``from each source dev set''. Mixing dataset and dev set might lead to questions otherwise.}
We then generate perturbations for evaluating each aspect of robustness by making use of seven perturbation methods and collecting human annotations. 

\subsection{Source Datasets}
\label{sec:source_dataset_section}
For building FREB-TQA, we leverage the development sets of four well-studied TQA datasets:  
WikiTableQuestions \citep[\textbf{WTQ},][]{Pasupat2015CompositionalSP}, 
\textbf{WikiSQL} \cite{Zhong2017Seq2SQLGS}, 
Sequential Question Answering \citep[\textbf{SQA},][]{Iyyer2017SearchbasedNS}, and 
Tabular And Textual dataset for Question Answering \citep[\textbf{TAT},][]{Zhu2021TATQAAQ}. 
The first three datasets feature tables from Wikipedia, TAT addresses the financial domain.
%\annetodo{confusing: here, you say that only TAT requires numerical reasoning, but isn't aggregation also some kind of numerical reasoning? the RQ question type also exists for WTQ?}
See \aref{source_dataset_overview} for detailed statistics.
For all source datasets, we eliminate questions that relate to
the table structure, such as \enquote{What is the name of the actor in the first row?}
To identify such questions, we use a word list provided in \aref{eliminate_position}.
Around 10\% of the questions in WTQ and 3\% of TAT questions are filtered out by this criterion.
%\heiketodo{Can we move this last paragraph to the end of Section 3.1? (and remove the phrase ``and question types'') It's not related to EQs and RQs and in particular not related to question type classification for TAT, right?}

\subsection{Extraction and Reasoning Questions}
\label{sec:qtc}
%MM:
%TQA involves identifying relevant cell values and, optionally, composing the answer by aggregating or comparing several values. 
%
% MM:
To decouple the robustness aspects in TQA, we group questions from the source datasets into extraction questions (EQs) and reasoning questions (RQs) \wei{by applying heuristics and classification models.} 
%\blue{in a semi-automatic manner} 
%\MMN{Not sure semi-automatic is specific enough. Make it more specific.}.
% \annetodo{not perfect, but not severe: up to here, this was always called "retrieval" in the paper}
The answer to an EQ can be retrieved from a single cell of the table.
The answer to an RQ additionally requires aggregating over several cell values. 
In \fref{fig:framework}, the question \enquote{How many floors does the Fosh Tower have?} is an EQ since it only requires retrieving a cell value.
%The question \enquote{How many floors does the Fosh Tower have?} in \fref{fig:framework} is an EQ, as it only requires identifying the correct cell.
The question \enquote{Who received the least amount of votes?} is an RQ as it requires comparing values of several cells.

\paragraph{Question type classification for WTQ.}
% AF: guide the reader more
WTQ provides a diverse set of questions. We group them into EQs and RQs \heike[as follows]{using two different methods}.
Each method is tuned to identify either EQs or RQs with high precision, as our final benchmark will only contain the set of questions identified as EQ or RQ by the respective method.\footnote{Around 20\% of questions are selected by neither of the models and thus eliminated from our benchmark. We also found that the final set of EQ and RQ cases do not overlap.} % AF: added this back, kind of crucial information, one reviewer also missed it

%\blue{We aim for a high-precision model for identifying each question type, as we only include the question labeled as EQ or RQ.}
%We build three models as \red{baselines}:\annetodo{why baselines?} 
%(1) lexical rule-based, 
%(2) LLaMA2, and
%(3) their combination. 
%MM:
Given a table, a question, and an answer, the \textbf{lexical rule-based} method conducts a string match between the answer and the table's cell values. % using word uni-gram. AF: unclear, not needed
If there is no %\wei[]{not}
match, the corresponding question is labeled as RQ.   
Otherwise, we detect if the question contains any comparative or superlative words using POS tags.\footnote{\href{https://www.nltk.org}{https://www.nltk.org}} 
Since these words are signals for aggregation over table cells, we mark the question as RQ. 
Otherwise, the question's type is set to EQ.
We also prompt \textbf{LLaMA2}-13b \cite{Touvron2023Llama2O} to label the question type (\heike{The prompt is provided in} Appendix \ref{llm_prompts}). 
Finally, we combine the lexical rule-based and LLaMA2 models. 
If the lexical rule-based model labels a question as EQ, we obtain the LLaMA's prediction in addition. 
If they agree, the question type is set to EQ. 
Otherwise, it is set to RQ.
To estimate the quality of these methods, we manually \heike[(\aref{annotations})]{} annotate questions sampled randomly from WTQ (100 EQs and 100 RQs). \heike{Details are given in  \aref{annotations}}.
% To estimate the quality of our automatic question classification method, we manually annotate questions sampled randomly from WTQ until we have obtained 100 EQs and 100 RQs.\footnote{The total annotated dataset comprises \textcolor{red}{100} EQs and \textcolor{red}{338} RQs.}
%We use precision to select the most precise method for identifying each question type. 

\begin{table}[!t] 
\footnotesize
\centering
  \begin{tabular}{@{}lcc@{}}
    \toprule  
    \textbf{Model}& \textbf{EQ}  & \textbf{RQ} \\
      \midrule
    LLaMA2 & 74.27& 57.02 \\
    Rule-based & 75.23 & \textbf{83.42}\\ 
    Combined & \textbf{93.81}& 60.17 \\
    \bottomrule
  \end{tabular}%
 \caption{The precision of examined models for question type classification on 200 questions from WTQ.} 
 %Best score for each question type is written in bold.}
  \label{model_selection}
\end{table}

Table \ref{model_selection} shows the results achieved by these models.
As we aim for a high-precision question-type classification, we apply the combined model for identifying EQs and the lexical rule-based method for identifying RQs on all questions in WTQ.
% Our dataset inspection shows that the models' predictions do not overlap. AF: what does this mean?

\textbf{Question type classification for WikiSQL.}
Questions in WikiSQL do not require complex reasoning \cite{Zhao2023RobuTAS,Lin2023AnIT}. 
%From WikiSQL, we select a set of questions that can be answered without \wei[executing any SQL operations]{aggregation operations} \wei[]{(annotation are provided in the dataset)} and that we label as EQs.
In WikiSQL, questions are labeled with regard to the operations required to answer them. We select the questions that can be answered without performing aggregation operations and label them as EQs.

\textbf{Question type classification for SQA.}
%SQA consists of a sequence of question-answers, posing a dialog about a table. 
The SQA dataset consists of dialogues in the form of questions and answers related to the information in a table.
The answers to all questions except for the first question in a dialog rely on the dialog history. 
Thus, we select only the first question from each sequence, which usually asks for retrieving information from the table \cite{Iyyer2017SearchbasedNS}. Hence, we mark them as EQs.

%\wei[]{. As the first question , we}{} marked \wei[]{them} as EQ.
%\annetodo{here I cannot follow, why is this always an EQ?}
%Similar to WikiSQL, SQA does not contribute to RQs in our benchmark. AF: not necessary to say that

\textbf{Question type classification for TAT.}
TAT consists of questions about tables and text snippets and requires numerical reasoning. %\heike[to be solved.]
The dataset also features annotations about how an answer is derived, whether the answer is based on the table or the text, and which operations are required to answer a question. We leverage these annotations to identify EQs and RQs. More specifically, we first select questions that can be answered only based on tables. Next, if a question needs derivation or comparison, we classify them as RQ. Otherwise, we classify them as EQ.
%\wei[]{The dataset also features annotations about how an answer is derived, where does the answer come from (table, text), if a question need comparison to be solved etc. We leverage the annotation to classify EQs and RQs. More specidically, we first selected questions that can be answered only from tables. Then, if a question needs derivation or comparison, we classify them as RQ. Otherwise, we classify them as EQ.}

%To identify these questions are annotated by human annotators in this dataset, indicating if they need any comparison or mathematical operations over table cells.   
%If a question needs comparison or mathematical operations, we identify it as RQ. \annetodo{super unclear, is the sentence about the human annotator the explanation? if yes please uncomment!!}
%Otherwise, it is set to EQ. 

%, suggesting the coverage for our dataset is not affected greatly.}
%So these questions are  unsuitable for robustness evaluation \cite{Zhao2023RobuTAS}.  
%\weitodo{not just related to tat, but ALL!}

%%% perturbations
\subsection{Perturbations for Testing Retrieval Robustness against Table Structure Changes}
\label{sec:tsr}
A robust TQA system should answer an EQ by retrieving the answer from a table, regardless of table structure changes.
We perturb the structure of tables associated with extraction questions. 
We replicate one perturbation type from previous work and introduce two new perturbation types for measuring this robustness aspect. 

\paragraph{Shuffle all rows (columns).}
Following \newcite{Zhao2023RobuTAS}, we randomly shuffle all rows (columns) in a table. 
This perturbation allows us to study if TQA systems are robust against changes in re-arranging all rows (columns).  
However, it does not reveal which biases may impact the robustness of TQA systems. 
Thus, we introduce the following two new perturbation methods for this aspect.  

\paragraph{Shift target rows (columns).}
%\paragraph{Target row front/middle/bottom .}
For each EQ, the \textit{target} row (column) in a table contains the cell that corresponds to the answer.
This perturbation type shifts target rows (columns) either to the top, to the middle, or to the bottom part of a table.  
We identify the target cell by applying exact match between the answer and table cell value.
%\annetodo{see before, what is a unigram string match exactly?  
For shifting target rows, we partition a table into three equal-length parts, referred to as top, middle, and bottom. 
For shifting target columns, we partition a table into two equal-length parts: front and back. 
We use more partitions for rows because on average tables include more rows than columns (Table \ref{perturbation_statistics}). 
For our benchmark, we remove the target rows (columns) from the table and re-insert them at a random position in each partition.
This perturbation method allows us to study \heike[if the position of the answer brings any bias to TQA systems.]{whether TQA systems exhibit any positional biases.} 

\paragraph{Transpose.} 
This perturbation type transposes the table, i.e., it rotates the table by 90 degrees and turns rows into columns and columns into rows.
\wei[\red{To remove the impact of header information on this perturbation, we transform headers to indices.}]{}
%\annetodo{what does this mean? pandas-style? \wei[]{yes, see the example I sent in the teaser image.}}
This perturbation allows us to study if TQA systems have a bias towards particular table layouts.

%\heiketodo{you are probably not done with it but just to be sure: if you rename it here, make sure to rename the second aspect everywhere, including the latex macro for it; I created another version of the macros with the first letter in upper case to be used in section titles}
\subsection{Perturbations for Testing Attention to Relevant Cells} % relevant cells
\label{sec:su}
TQA systems may answer a question by exploiting shortcuts (model-internal knowledge or positional biases) without paying attention to relevant cells (cells that are important to compose an answer). % (as shown by \newcite{} for Table NLI).
We propose three new perturbation methods for reasoning questions (RQs) to study this aspect of TQA robustness. 

\paragraph{Remove relevant cells.}  
We perturb a table by removing relevant cells from tables. 
The relevant cell annotations associated with RQs in our benchmark have been created by \citet{Zhu2021TATQAAQ} and \citet{Ye2023LargeLM} for TAT and WTQ, respectively. 
We test the validity of the latter since the annotations are gathered from LLMs (see \aref{annotations}).
%Since the relevant cells annotations from \citet{Ye2023LargeLM} are obtained via large language models. 
%We test its validity before using them. Details can be found in Appendix \ref{annotations}.
This perturbation lets us investigate to what extent TQA systems bypass relevant cells to derive their answers.
We observe that 70\% of relevant cells contain
non-numerical values for WTQ. For TAT, all relevant cells contain numerical values.

\paragraph{Remove table.}
TQA systems may bypass the whole table and use their internal knowledge to answer a question.
To test for this behavior, for any RQs, we replace the table with a dummy table, consisting of one cell with a \enquote{None} value.  

\paragraph{Shift relevant rows.}
This perturbation evaluates to what extent TQA systems bypass table cell values and rely on the position of relevant cells. 
%Since a cell position is defined by its row and column indices, we focus on rows that contain relevant cells. 
For instance, to answer the question \enquote{Who received the least amount of votes?} in Figure \ref{fig:framework}, a TQA system may exploit a shortcut between the last row and the question  since in most cases, rows in tables from TQA datasets are sorted.  
Because of the correlation between cell values and positions, this type of shortcut mostly occurs for RQs that require comparing cell values.
%\red{RQs that need \wei[aggregation of cell values] {comparing cell values}.}
%\annetodo{definition of aggregation: "An aggregation is a collection, or the gathering of things together" -- i.e., comparison is not aggregation! "aggregation among" (was here before) does not work either. Better: just use "comparison"}
%For any of these questions,
In this perturbation type, for each RQ, we remove relevant rows \wei[from the table]{}and re-insert them at a random position of the table. 
%One distinctive feature between this positional bias and the ones we discuss in Section \ref{sec:tsr} is the former comes from unwanted linkages between questions and cell positions while the latter comes from cell positions purely. 

%\subsection{Perturbations for \blue{Testing} Aggregation Robustness against Numeric Value Changes} %AF: I have a kind of garden path issue when reading this section title, suggestion:

\begin{figure}[!t]
    \centering
    \includegraphics[width=0.9\columnwidth]{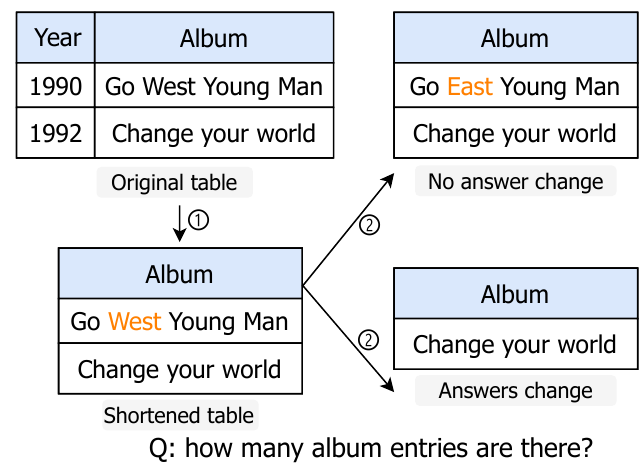}
    \caption{
   An example of aggregation/comparison robustness in case of string change. \ding{172} illustrates shortening an original table to table cells on which numerical aggregations or comparisons operate. \ding{173} illustrates modifications on a shortened table, leading either to a change in answer or not. The orange parts mark changed strings.
   %\MMN{change the font family of the text in the table. It should be consistent with rest of the paper. Also valid for Figure 1. }
   }
    \label{fig:string_change}
\end{figure}

\subsection{Perturbations for Testing Robustness when Aggregating/Comparing\wei[Numeric]{} Values} % keep it simple(r) :)
\label{sec:nrr}
TQA systems should compose a correct answer to an RQ by aggregating \heike{over} or comparing values in the provided table
independent of value changes. 
%\heiketodo{I don't think that ``in response to'' is the correct phrase here. Better: ``independent of value changes'' or ``even in the presence of value changes''}
%rather than leveraging model-internal knowledge or positional biases.}
%\wei[]{/comparing} \st{numeric} values in a table. 
% AF: was "aggregating  over": (to) aggregate sth. -- does not need a preposition!!
To evaluate this capability, for RQs, we manually modify cell values that should be aggregated to answer questions.
We ask three human annotators to first identify the table cells that numerical aggregations or comparisons operate on \wei[]{(step \ding{172} in Figure \ref{fig:string_change})}.
%\annetodo{feel free to change this back after adding an example} 
We refer to the part of the table relevant to answering the question as the \textit{shortened table}.
We use the shortened table versions in our experiments to minimize the effect of table length in this test.
Then, annotators are asked to change one or two cell values per question, once resulting in an answer change and once not resulting in an answer change \wei[]{(step \ding{173} in Figure \ref{fig:string_change})}.
We define two types of perturbations.

\noindent\textbf{Modify values to change answers.}  % 
%We modify cell values to change the answers, accordingly. 
This perturbation allows us to study to what extent TQA systems perform correct aggregations and adapt answers to the value changes accordingly.  
An example of numeric value changes is shown in Figure \ref{fig:framework}. 
The votes for \textit{Olsson} is changed from \textit{4} to \textit{361}, resulting in a change of the answer (from \textit{Olsson} to \textit{Leslie}). \wei[]{Figure \ref{fig:string_change} shows an example of string value changes. The first entry \textit{Go West Young Man} is removed, changing the answer from \textit{2} to \textit{1}}.
%, who now has most votes (\textit{15}). 
%Figure \ref{fig:string_change} shows an-other example, where value changes happen in the context of strings: West is changed to East without changing the answer (2).% AF: extended explanation a bit so it also works without actually going back to Fig. 1
%\red{add explanation for the counting example with string changes}.

\begin{table}[!t]
\small
\centering
%\resizebox{\columnwidth}{!}{%
  \begin{tabular}{@{}lcccccc@{}}
    \toprule
    \textbf{Aspect}&
      \textbf{\#P} &  \textbf{\#Q} &  \textbf{\#QT}&  \textbf{\#R}&  \textbf{\#C}&  \textbf{\#A}\\
    \midrule
    RR-TSC & 64890 &6489&10.05&12.14&7.18&1.02\\
    Rel-Cel &6303&2101&14.75&21.44&5.77&1.05\\
    ACR-VC & 4012&2006&14.62&21.70&5.61&1.05\\
    \midrule
    Total & 75205 &8590&10.28&14.42&6.85&1.03\\
    \bottomrule
  \end{tabular}
%}
\caption{
Benchmark statistics grouped by robustness aspects: 
retrieval robustness in case of table structure changes (RR-TSC), attention to relevant cells (Rel-Cel), and aggregation/comparison robustness in case of value changes (ACR-VC). % AF: see comment in that section, can we rephrase this a bit?
\#P and \#Q show the number of perturbations and questions, respectively.  
\# QT, \#R, \#C, and \#A show the average question length in tokens, the number of table rows, the number of table columns and the number of cells for composing an answer. 
}
\label{perturbation_statistics}
\end{table}

\noindent\textbf{Modify values without changing answers.} %
%We modify cell values so that the answer does change.
This perturbation aims to study if systems articulate correct answers because of their biases to certain values. 
For instance, in Figure \ref{fig:framework}, systems might be capable of comparing \textit{15} and \textit{4}. 
However, if we change the votes for \textit{Leslie} from \textit{15} to \textit{1500}, without changing the answer, systems might face difficulties as \textit{1500} might not fall into the cell values distribution during pre-training. 
\wei[]{Figure \ref{fig:string_change} shows an example of string value changes: \textit{West} is changed to \textit{East} without changing the answer (2).} %\heiketodo{In the figure caption: are you sure, the ordering is correct? shortened table is the very last one, right? but this does not fit to the description}

%To create data for both types of perturbations, 
%We only change one or two cells to reduce the workload of annotators while still obtaining interesting perturbations.   
We filter out instances that annotators find unanswerable.
In the case of TAT, all modified values are numeric. In the case of WTQ, 50\% of the modified cell values are numeric, the others are string-based.
%\wei[]{Most modified values are numeric (approx 50\% for WTQ\footnote{The number is different from the 70\% presented in the previous section because we calculate the percentage among the modified cells by annotators here, rather than among the relevant cells.}and 100\% for TAT).}

To assess the quality of the output of these perturbations, for each perturbation type, we randomly sample 50 instances created using this type from each dataset, resulting in 200 instances in total.
Then, we ask two annotators who were not involved in the perturbation creation to provide the answers given the changed tables and questions.
We compute the exact match accuracy of answers provided by the two annotators, which amount to 92.5\% and 93.5\%.
We find that most wrong cases are related to questions asking for percentage changes: here, annotators sometimes neglected to add the minus symbol to negative percentage changes.

  \section{Experimental Settings}
\label{sec:exp}
We use our benchmark to evaluate the robustness of state-of-the-art TQA systems with regard \heike[to different]{to our proposed} aspects. 
Table \ref{perturbation_statistics} shows the main statistics of our benchmark grouped by the robustness aspects for which TQA systems are evaluated.

\subsection{Examined TQA Systems}
\label{sec:system}
%We evaluate \textcolor{red}{the following models}
%\annetodo{and why were they selected?}:\annetodo{before, the paper talks about 2 types of models - needs to be adapted! Tapas and TAPEX are also based on LLMs, right? GPT-3.5 is not fine-tuned, right, but LLaMA is? -- unclear atm} TaPas \citep{Herzig2020TaPasWS} TAPEX \citep{Liu2021TAPEXTP}, OmniTab \citep{Jiang2022OmniTabPW}; GPT-3.5-turbo and LLaMA2 fine-tuned with LoRA \cite{hu2021lora} and Binder \cite{Cheng2022BindingLM}. 
%\textcolor{red}{We choose them because they feature a common and good collection of the current approaches in TQA, leading to SoTA performance.  \MMN{"good is not a precise reason to select a model."}
%TaPas, TAPEX and OmniTab are pre-traiend end-to-end TQA systems. 
%GPT-3.5 and LLaMA are LLMs and Binder combines LLMs with symbolic languages.} 
%\heiketodo{clearly state that these models cover different types of TQA systems and therefore represent current state of the art (pipeline, end-to-end) and which one is which}
We analyze the robustness of three types of systems: end-to-end systems that are fine-tuned for the TQA task; pipeline systems that generate relevant cells or SQL queries which are then executed on a table, and off-the-shelf LLMs.
%\annetodo{pleaes check: I think "large language model" may occur earlier in the paper and the acronym should be introduced at that point} 
In particular, we compare the following TQA systems.
\textbf{TAPEX} \cite{Liu2021TAPEXTP} is an end-to-end TQA system based on BART \cite{Lewis2019BARTDS}.
\textbf{OmniTab} \cite{Jiang2022OmniTabPW} 
further fine-tunes TAPEX on both more natural and synthetic data. %\heiketodo{on both more natural and synthetic data ? (typically, ``natural'' and ``synthetic'' is a contradiction, so we need to add ``both'' or something like that}
\textbf{TaPas} \cite{Herzig2020TaPasWS} first predicts relevant cells and an aggregation function, backboned by BERT \cite{Devlin2019BERTPO}. 
Then, it articulates answers based on outputs from the previous step with a numeric tool. We categorize it as a pipeline model as answers are not directly generated.
\textbf{Binder} \cite{Cheng2022BindingLM} is a pipeline model consisting of a parsing and an executing step. 
First, intermediate representations (e.g., SQL queries) are generated by GPT-3.5, and then the queries are executed by a program interpreter.
\textbf{GPT-3.5} is an LLM and the backbone of various TQA models \cite{Ye2023LargeLM,Zhang2023ReAcTableER}.
%\annetodo{here, an explanation is lacking how it's used: zero-shot?} 
In the prompt to GPT-3.5, we use a three-shot demonstration to obtain answers (see \aref{prompt_gpt_llama}).
%\annetodo{it is a bit unclear why that is in the appendix -- should we somehow clarify that this has been added later? or mention this somehow elsewhere? any ideas, Mohsen, Heike?}}
LLaMA \cite{Touvron2023Llama2O} is an open-source LLM. 
We fine-tune its 7b chat version with LoRA \cite{hu2021lora}.
We describe details of fine-tuning in Appendix \ref{train_prompt_details}. 
%\MMN{we don't need appendix here. As I suggested in the related work: bring the explanation of those models here. When I looked at the tables without reading the results section, it's not clear what is LLMs and what is end-to-end? What happend to pipeline approaches introduced at the beginning of the paper? }
%\MM[]{
%We investigate the robustness of the following TQA systems since they represent current state of the art:
%\paragraph{TaPas.} ...  
%\paragraph{TAPEX.} ... 
%\paragraph{OmniTab.} ... 
%\paragraph{GPT-3.5.} ... 
%\paragraph{LLaMA.} ... 
%\paragraph{Binder.} ... 
%}
%\input{latex/sometable}

% Note that the systems we evaluate feature different input length.
% For fair comparisons among systems, unless explicitly mentioned, results are calculated based on only examples with length smaller than 512 after tokenization.\footnote{This reduces the size of the benchmark only little as more than 90\% of examples have shorter length.} 

%\annetodo{this confuses me: doesn't this mean that also for some original instances, the relevant cell may or may not be within in the input length restriction? are most tables this short? -- discuss in meeting?}
%\heiketodo{can you give a percentage how much this actually is? i.e., the benchmark still 90\% of its size or rather 20\%?}
%\textcolor{red}{all above 90\%}
%Details can be found in the Appendix.

\subsection{Evaluation Metrics}
To compare results on our benchmark, we use the following metrics.

\noindent\textbf{Exact match accuracy (Em)} checks if the predicted answers and the ground truth are the same.
It is a widely used metric for evaluating TQA systems \citep{Pasupat2015CompositionalSP,Yang2022TableFormerRT,Jiang2022OmniTabPW}. 

\noindent\textbf{Exact match difference} \citep[\textbf{Emd},][]{Zhao2023RobuTAS} measures system performance change before and after perturbations (negative values indicate performances drop). 
%\citet{Zhao2023RobuTAS} use it to analyze robustness of TQA systems.
Both Em and Emd focus on overall system performance.

\noindent\textbf{Variation percentage} \citep[\textbf{VP},][]{Yang2022TableFormerRT} measures to what extent predictions change before and after performing perturbations from an instance-level perspective. 
%\MM[It  is calculated as follows, where t and f stand for correct and wrong predictions respectively. 
%(t2f means pre-perturb correctly predicted and post-perturb incorrectly predicted).
%\begin{equation}
%    V P = \frac{(t2f + f2t)}{(t2t + t2f + f2t + f2f) }
%\end{equation}
%]{
It is defined as follows: 

\begin{equation}
 \text{VP} = \frac{\text{C2W} + \text{W2C}}{\text{N}}
\end{equation}
where $\text{C2W}$ counts the number of instances whose predictions change from correct to wrong and $\text{W2C}$ is the number of instances whose predictions change from wrong to correct.
N is the total number of instances. 
\wei[To reduce the effect of randomness in experiments, for perturbations involving randomness,]{For perturbations involving randomness,}
%\heiketodo{This sentence is very complicated. Much easier: ``For perturbations involving randomness, we report ...'' (omit this whole ``to reduce...'' clause)} %(e.g., random shuffle), % AF not needed, this is clear
we report the mean and standard deviation of scores over five runs with different random seeds.

   \section{Experimental Results} 
\label{sec:result_tsc}
In this section, we discuss the performance of TQA systems on our benchmark, and provide a detailed analysis of these models for each robustness aspect.

\subsection{\Firstaspect}

\begin{figure*}[!t]
    \centering
    \includegraphics[width=2\columnwidth]{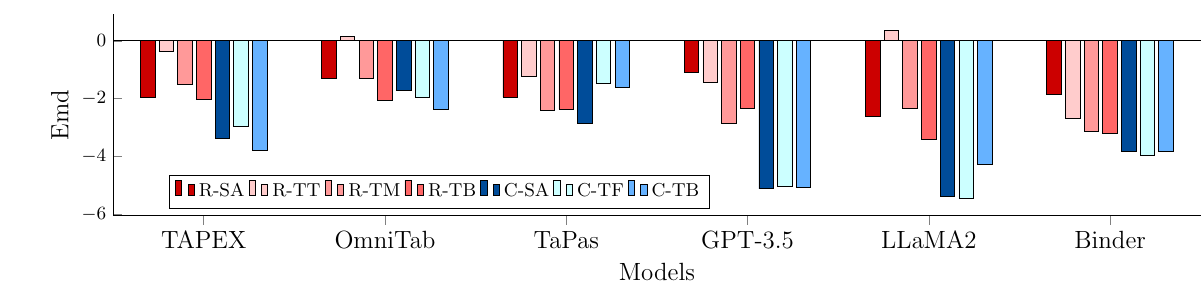}

    \caption{Exact match difference (Emd) on retrieval robustness against table structure changes perturbations \textbf{for extraction questions}, averaged across four datasets and seeds. \textsc{R}, \textsc{C} and stand for row and column. \textsc{SA}, \textsc{TT},\textsc{TM}, \textsc{TB} and \textsc{TF} stand for shuffle all, target top, target middle, target bottom/back, target front, respectively.
    %\MMN{The font family of the model names and should be identical with the style of the text. Why are the model names not written horizontally? Do it for all figures. }
    }
    \label{fig:result_tsc_emd}
\end{figure*}

\begin{figure*}[!t]
    \centering
    \includegraphics[width=2\columnwidth]{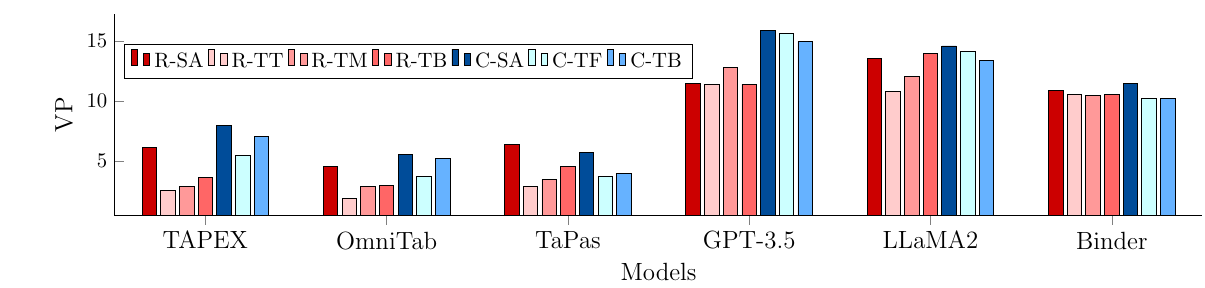}

    \caption{Variation percentage (VP) on retrieval robustness against table structure changes perturbations \textbf{for extraction questions}, averaged across four datasets and seeds. \textsc{R}, \textsc{C} and stand for row and column. \textsc{SA}, \textsc{TT},\textsc{TM}, \textsc{TB} and \textsc{TF} stand for shuffle all, target top, target middle, target bottom/back, target front, respectively.}
    %\MMN{in all figures and tables, if italic is used in the caption, it should be used in the table and figure too.}}
    \label{fig:result_tsc_vp}
\end{figure*}

%\annetodo{Q1 earlier vs. RQ1: should be consistent}

To study this aspect of robustness, we use the first part of our benchmark (see Section \ref{sec:tsr}) which consists of shuffling rows (columns), shifting target rows (columns), and transposing tables for EQs.
To rule out the effect of different maximum input lengths of the models in this test, we use only instances that are within the maximum input length (512) of TaPas, the model accepting the smallest input. This means we make use of 91\% of the EQs.

Figure \ref{fig:result_tsc_emd} and Figure \ref{fig:result_tsc_vp} show Emd and VP scores for row and column shuffling, respectively, averaged across all source datasets.
We report the detailed results for each dataset in Appendix \ref{appendix_tsc_results}.
In addition, the first column of \tref{tab:results_tsc_tr} provides the model performance in terms of Em on the original data (without perturbations). 

Figure \ref{fig:result_tsc_emd} shows that the performance of almost all systems drops when evaluating them on row (column) shuffling, which is consistent with prior results \cite{Zhao2023RobuTAS}. 
However, our fine-grained benchmark enables us to draw more conclusions beyond this general observation.
First, systems are more vulnerable to column shuffling than to row shuffling. 
This is apparent for LLMs (GPT-3.5 and LLaMA2). 
When comparing model types, LLMs are most affected by row and column shuffling, followed by pipeline systems. 
The end-to-end TQA systems (TAPEX and OmniTab) are more robust in this regard. 
%OmniTab is the most robust system both from the perspective of performance drop (Emd) and instance-level variation (VP).
%
Second, regarding row perturbations, systems are highly impacted by the position of the target row.
The more \heike[we move it]{it is moved} down the table (TM and TB), the more the performance drops. 
Notably, moving the target row to the top (TT) even leads to performance improvements for some of the systems,  confirming that the systems have encoded some positional biases.
%\annetodo{how does this related to max. input length?  below there is sth. on maximum input window, but it might be good to say how this is in general with the systems in the study} 
It further reveals that systems are likely to fail on tables with many rows, where the target row can be located anywhere. 
Similar observations are found by  \citet{Lin2023AnIT}. 
 %\annetodo{what about the RQs? or do you mean "selected questions"?}}\annetodo{is this true only for this experiment or for all of them?}
However, we show that even within \wei[models' maximum input window]{the maximum input window of a model}, %\heiketodo{``within the maximum input window of a model'' sounds better} 
the positional bias exists during cell retrieval.

\begin{table}[t]
\small
\centering
%\footnotesize
  \begin{tabular}{lcccc}
    \toprule  
    \textbf{Model}& \textbf{Original Data}&\textbf{Emd}&\textbf{VP}&\textbf{Emd-Ft}\\
    \midrule
     TAPEX  & 79.83&-55.36&59.64&-43.11 \\
      %\midrule
     OmniTab  & \textbf{82.24}&-57.84 &60.21&-41.71\\
      %\midrule
     TaPas  & 71.82&-60.98 &65.59&-46.83\\
      %\midrule
     Binder   & 67.55&-51.86&62.70&-\\
     GPT-3.5  & 71.77 &\textbf{-17.32}&\textbf{26.15}&-\\
      %\midrule
     LLaMA2  & 61.10 &-30.58&38.42&-\\
      %\midrule
\bottomrule  
\end{tabular}%
 \caption{Exact match difference (Emd) and variation percentage (VP) for \textbf{table transposing}.
% \textit{ORI} stands for original performance without any perturbations. 
 Emd-Ft stands for Emd after fine-tuned with transposed tables.
 Bold values suggest best performances.
 %\MMN{Are these explained in the experimental setup?}
 %\MMN{Do not use unnecessary midrules. I corrected it here as a role model. Please check the other tables yourself.}
 }
  \label{tab:results_tsc_tr}
\end{table}

Third, \tref{tab:results_tsc_tr} shows that systems fail on transposed tables as well (see columns Emd and VP). Only GPT-3.5 suffers less. We assume that this model has seen a larger variety of tables during its pre-training.
Since our analysis of table transposing is new and especially the fine-tuned end-to-end TQA systems perform considerably worse on that perturbed data, we look into this aspect in more detail and investigate their ability to adapt to new table structures (here: transposed tables).
For this, we further fine-tune them on transposed table data.
The last column of Table \ref{tab:results_tsc_tr} shows the results. 
By fine-tuning on transposed data, the performance of these models improves.
However, the gap to GPT-3.5 is still large, showing TQA systems \wei[]{might} need architectural changes or pre-training datasets that feature more diverse table structures. %\heiketodo{in my opinion, this is phrased too strongly given that we are only speculating. We could say ``might need'' to hedge it a bit.}

\subsection{\Secaspect}
\label{sec:result_shortcuts}
To investigate if systems use relevant cell values to compose answers or rather use shortcuts, such as implicit knowledge or biases, we use the second part of our benchmark (see Section \ref{sec:su}). 
To the best of our knowledge, this aspect has never been tested in prior work on TQA robustness.
We split our analysis \heike[in]{into} two parts. 
First, we investigate system behavior when removing relevant information (relevant cells or the whole table). 
Second, we \heike[look into]{examine} to what extent systems exploit linkages between questions and relevant cell positions rather than paying attention to cell content.
%\red{unwanted}\annetodo{can we say potentially unwanted? all machine learning ultimately relies on biases ... bias it not necessarily "bad". unreliable? potentially undesired? potentially harmful? (I prefer the latter) \wei[]{that's valid point, it depends on how one view bias. Reviewer 2 has some points about leveraging biases to improve task performance, but we emphasize in our study, biaes is generally "bad" because we focus more on robustness rather than task performance in this study.}} linkages between questions and relevant cell positions.
%\weitodo{why this is unwanted, expand it a little bit. Linkages between cell positon and performance could potentially harm robustness}

\begin{table}[!t]
%\large
\small
\centering
%\resizebox{\columnwidth}{!}{%
  \begin{tabular}{@{}lcccccc@{}}
    \toprule
     \multirow{2}*{\textbf{Model}}  & \multicolumn{3}{c@{}}{\textbf{WTQ}} &
  \multicolumn{3}{c@{}}{\textbf{TAT}}   \\
\cmidrule(lr){2-4}  \cmidrule(lr){5-7}
      & \textbf{ORI} & \textbf{RT} & \textbf{RRel} & \textbf{ORI} & \textbf{RT} & \textbf{RRel}  \\
    \midrule
    TAPEX* & 52.26& 8.84 & 9.57 & 9.65&3.53&5.38\\
    OmniTab* & \textbf{58.12}& 7.76 & 12.55 & 10.02&3.90& 6.12\\
    TaPas** & 47.02& \textbf{2.80} & \textbf{8.21}&-&-&-\\
    Binder** & 45.67 &6.86&11.62& \textbf{37.65}&\textbf{0}&\textbf{1.01} \\
    GPT-3.5&42.06& 6.95 & 16.52 &36.17&\textbf{0}&1.73 \\
    LLaMA2 &35.92& 7.44 & 11.91& -&-&- \\
    \bottomrule
  \end{tabular}
%}
\caption{The exact match (Em) of examined TQA systems regarding attention to relevant cells. 
ORI shows results on original tables. 
RT and RRel show the results when the removing table perturbation and  removing relevant cells are used, respectively. 
We do not report results for TaPas and LLaMA2 on perturbations for the TAT subset as their ORI accuracy scores are too low (<4\%) to \heike[derive concrete analysis from]{draw conclusions from them}.
}
\label{results_shortcut_1}
\end{table}

\tref{results_shortcut_1} shows the Em results for the first part, analysing the behavior of systems when removing relevant information. 
The column ORI shows the original performance of systems, the column RRel shows the performance when removing relevant cells and the column RT shows results when removing the whole table and replacing it with a dummy table. 
The results indicate that TaPas pays most attention to the cell values, i.e., its Em score is the lowest among the systems when removing the relevant information, which is the desired behavior.
%\wei[performance drops the most]{Em score being the lowest (desired behaviour)} when removing relevant information.

In general, the pipeline models are more robust against changes \heike[to the relevant cell position]{in the position of the relevant cells} than end-to-end models.
This intuitively makes sense as they consist of an executing component that performs a function or query directly on the structured table. 
For a deeper investigation, we analyze instances which the end-to-end systems TAPEX and OmniTab still predict correctly after removing relevant information in tables for TAT.
All of them feature the same answer ``2019'', indicating that answer values are not well distributed in the TQA benchmarks and systems learn this bias during fine-tuning.

For investigating the second question, i.e., to what extent systems exploit positional biases when answering questions, we \heike[base our analysis on] {analyze} the variation percentage (VP) with and without shuffling relevant rows or columns. 
To account for the effect that shuffling data leads to challenges on its own (the first robustness aspect we analyzed before), we compare VP for comparison questions (involving cues, such as \enquote{least}) with VP for non-comparison questions. Table \ref{relevant_shortcut} shows the results. For all systems, the gaps are larger than zero, indicating that they all use question-related shortcuts. 
Among them, TaPas features the largest gap, suggesting that among the tested systems, it exploits the most question-related shortcuts. 

\begin{table}[!t]
\small
\centering
%\footnotesize
  \begin{tabular}{lccc}
    \toprule  
    \textbf{Model}& \textbf{Compare}&\textbf{Non\_compare}&\textbf{Gap}\\
    \midrule
     TAPEX  & 10.24&9.70&0.54\\
      %\midrule
     OmniTab  & 7.37&6.02 &1.35\\
      %\midrule
     TaPas  & 10.98&\textbf{4.04} &6.94\\
     Binder & \textbf{9.47}&9.07&0.40\\
      %\midrule
     GPT-3.5  & 16.62 &16.40&\textbf{0.22}\\
      %\midrule
     LLaMA2  & 17.11 &12.89&4.22\\
      %\midrule
\bottomrule  
\end{tabular}%
 \caption{VP of instances (not) requiring value comparisons %\red{comparing to be solved}
 on WTQ. \textbf{Compare} refers to instances requiring comparison of cells to be solved (e.g., questions contain \enquote{highest}, \enquote{lowest}) and \textbf{Non\_compare} refers to instances do not require comparisons. \textbf{Gap} refers to the gap between the two settings.}
  \label{relevant_shortcut}
\end{table}

\subsection{\Thirdaspect}

\begin{table*}[!ht]
\centering
%\LARGE
\small
%\resizebox{\columnwidth}{!}{%
  \begin{tabular}{lcccccccc}
    \toprule
     \multirow{2}*{\textbf{Model}}  & \multicolumn{4}{c}{\textbf{WTQ}} &
  \multicolumn{4}{c}{\textbf{TAT}}  \\
\cmidrule(lr){2-5}  \cmidrule(lr){6-9}
      & \textbf{ORI} & \textbf{ST} & \textbf{AC-VP} & \textbf{NC-VP} & \textbf{ORI} & \textbf{ST}  & \textbf{AC-VP} & \textbf{NC-VP} \\
    \midrule
   TAPEX & 47.55 &  48.03&  30.00 &  5.65 &  9.68& 10.17& 8.38& 2.23\\
    OmniTab &\textbf{52.79}& \textbf{54.69}& 26.46 & 4.76 & 10.06&12.96& \textbf{6.33}&\textbf{2.05}\\
    TaPas &41.09& 48.57& 16.39 &\textbf{3.81} &-&-&-&-\\
     Binder&42.99&50.88 &\textbf{14.56}&7.96& \textbf{37.78}&\textbf{41.19}&9.50&8.19 \\
    GPT-3.5&38.91&51.70& 23.20 & 8.78 &36.30&46.37&26.82&14.71 \\
    LLAMA2 &32.59&33.88& 32.45 & 6.31& -&-&-&- \\
    \bottomrule
  \end{tabular}
%}
\caption{System evaluation on aggregation/comparison robustness in case of value changes. ORI stands for original performance without perturbations. ST stands for passing short tables on which numerical aggregations operate. We report Em for the ORI and ST settings. AC-VP and NC-VP stand for variation percentage for the answers change and not change settings. We do not report results for TaPas and LLaMA2 on the TAT subset, as the performances are too low (<4\%) to derive reasonable analysis from. Bold values suggest best performances.}
\label{numerical_reasoning_short}
\end{table*}

We investigate the numerical reasoning abilities of TQA systems, i.e., their robustness against value changes. 
As many tables contain numeric data and questions might require the aggregation of several numeric values, this aspect is of utmost importance \heike[in the real world]{for real-world applications}. 
While numerical reasoning abilities of models have been analyzed for other domains or tasks \cite{Stolfo2022ACF,Akhtar2023ExploringTN}, a targeted analysis for TQA is missing in prior work. 
We analyze this aspect using the third part of our benchmark (Section \ref{sec:nrr}).
Table \ref{numerical_reasoning_short} shows the Em results for tables with original values (ORI) and shortened tables (ST) that contain only the part of the cell necessary for composing the answer.
%\annetodo{I feel the shortened table aspect needs to be clarified in the dataset construction Section}  
%on which aggregations operate.%\annetodo{on which? \wei[]{shortened tables}}
%\annetodo{TODO: make the intuition clear here again}
Additionally, the VP between the results before modifying the cell values and those after changing values is provided, once for the case where answers change (AC) and once where they do not (NC).
%\annetodo{Heike, Mohsen, is this clear?} \heiketodo{I think, yes} \MMN{Yes, this is. }
%\annetodo{this sentence is a bit long: break into the points it wants to convey}
For shortened tables, performance increases compared to original tables for all systems, again indicating that systems might perform a decent job on tables with a few rows but strongly fail on tables with many rows.
%Thus, we base our experiments on \st{numeric} value changes on shortened tables to rule out the effect \wei[of]{that} table length \wei[]{brings during the retrieval}.\annetodo{need to say this earlier}
%\annetodo{is this just about cell identification/retrieval/extraction? if yes: say that explicitly using the terminology used earlier in the paper} 
%\weitodo{I would like to delete this sentence, because the logic "thus" sounds a bit weird to me.}
When changing values, the pipeline systems TaPas and Binder are among the most robust for both AC and NC settings. 
We account this to the fact that they execute a predicted function or query on the table and, thus, directly involve the cell values when deriving the final answer.
End-to-end systems (TAPEX and OmniTab) show small VP on TAT.
However, they do not outperform LLMs, e.g., GPT-3.5. % \st{and Binder}.

\section{Discussion}
%\wei[]{We summarize the main take away message for this paper by building a profile for each approach, as described above. We suggest building more robust TQA systems by taking the pipeline approach. }

%\annetodo{Comment by reviewer: If you plan to submit this as the long paper, instead of writing Chapter 6 with a simple listing of facts and yourspeculations, accompany it with an empirical analysis of the experimental results for your speculations. Discuss indetail the reason for your experimental results, their significance, and how they can be applied to future research. -- I kind of agree, if we have space, as I say below: merging Discussion and Conclusion may be an idea, too}
\MM[In our experiments using our new benchmark FREB-TQA, we have analyzed end-to-end, pipeline, and LLM-based TQA systems.]{
In our experiments, we use our new FREB-TQA benchmark to analyze end-to-end, pipeline, and LLM-based TQA systems.
}
%AF: the below feels repetitive
%we analyze different state-of-the-art TQA systems on three fine-grained aspects of TQA robustness. 
%In particular, we investigate two end-to-end TQA systems, two pipeline systems and two LLMs.
Our experimental results show that all examined systems suffer from substantial issues when it comes to robustness.
However, different system types show different patterns. 
\textbf{End-to-end TQA systems, for instance, seem to be more robust against changes in the \wei[structure of the table]{row/column arrangement.} However, they are more likely to fail on numerical reasoning questions.} This might be because the datasets these models are pre-trained on do not feature complex questions requiring numerical reasoning.
%\annetodo{reviewer would have liked to see a cause mentioned - do we have anything here?}

\textbf{LLMs are more affected by row or column perturbations but much more robust against transposing tables.}
Their performance in performing aggregations or comparisons is highly dependent on the \heike{length of the serialized tables}, i.e., they perform much better on \heike[shorter tables]{tables that can be serialized to fewer tokens}. 
%\MMN{Paper length in terms of number of rows or columns?}
%\weitodo{Mohsen: Paper length in terms of number of rows or columns?; Wei: I think table length already suggests it is the number of rows or columns.}
%\heiketodo{Is the new version fine? I think it's clearer if we mention ``serialized'' again here}

%\annetodo{need to answer: input length!}

\textbf{Finally, the pipeline models in our study are more robust against changes in relevant cells, including value changes.} This is likely due to their symbolic execution component, which executes a predicted function or query on the given structured table.
However, the prediction of the function or query itself still suffers from various perturbations of table data, including changes in table structure.
Yet, another benefit of pipeline TQA systems is that the intermediate representation, i.e., the predicted function or query, makes the model explainable to a certain extent.
We hence argue that \textbf{more research in the pipeline-based paradigm is a promising step towards more robust TQA.}

%\red{Our analysis has revealed their advantages and shortcomings, which pave the way for future research on the robustness of table question answering systems.}
%\annetodo{better: finish with a meaningful sentence: what exactly have we found? I think it might actually make sense to end with a single section "Discussion and Conclusion", merging the two sections. The Conclusion feels a bit "empty" atm}

\section{Conclusion and Outlook}

In this paper, we have proposed to evaluate three aspects of robustness of TQA systems: retrieval robustness in case of table structure changes, attention to relevant cells, and aggregation/\heike[comparing]{comparison} robustness in case of value changes.
We have presented a novel benchmark for a fine-grained analysis of those aspects.  
The main building blocks of our benchmark are targeted table perturbation methods and high-quality human annotations.
Finally, we have evaluated a range of architecturally varied state-of-the-art TQA systems, as well as off-the-shelf LLMs.
Our study has shown that while none of the systems was consistently robust, their weaknesses and strengths differ from each other. 
Systems trained in an end-to-end fashion are able to deal with changes in \wei[table structure]{row/column arrangement}\MM[,]{} but LLMs perform better when it comes to numeric operations. 
The answers of pipeline-based models, which use symbolic methods for part of their computations, are more faithful towards the table they are supposed to use when composing their output. 
%As pipeline systems offer at least some explainability, we argue that research efforts should concentrate on these types of systems.

%\green{Our new benchmark constitutes an important first step toward comparing TQA systems in a finer-grained way, thereby directing research efforts towards building more robust TQA systems.}
%\wei[]{Future work could explore the possibility of utilizing LLMs in a pipeline manner to build more robust TQA systems.}
%\red{Here, suggest 2-3 steps for future work: what would you do for your next benchmark? how would you concrete improve a system? etc.}
%\green{Current TQA systems are less robust for long tables. Aggregating or extracting information from long tables is an important next step, and further benchmarks and analyses are needed.}

%\heiketodo{The last part of the conclusion is a bit confusing: jumps back and forth from findings, future work and contributions. I made a more structured suggestion below: see commented tex}

\heike{Our new benchmark constitutes an important first step for evaluating TQA systems in a fine-grained way, thereby directing research efforts towards building more robust TQA systems. As pipeline systems offer at least some explainability, we argue that research should concentrate on these types of systems. In particular, future work could explore the possibility of utilizing LLMs in a pipeline manner to build more robust TQA systems. Further, aggregating or extracting information from long tables is an important next step given the shortcomings of current systems in this regard which our analysis revealed.}

\section*{Acknowledgments} 
We thank the \heike[]{anonymous} reviewers for \heike[]{their} helpful feedback. We also thank Tianxing Liu, Xinyue Shi and Yidan Chen for their annotations. 
%\heiketodo{Are they fine with being mentioned by their names?}

% We thank PatBase, MineSoft and RWS for giving
% us permission to release the patents along with their
% metadata and automatic translations. We also thank
% Ulrich Klingner for creating the annotations on the
% InjVal part of the dataset and for his valuable ex-
% planations. We also thank Irene Kitsara and Patrick
% Fievet from the World Intellectual Property Orga-
% nization (WIPO) for an insightful discussion and
% information on the WIPO-related patent landscape
% studies. We also thank Tim Tarsi for fruitful discus-
% sions.
 %\input{latex/7-conclusion}
% \input{latex/responsibleNLPresearch}
\section*{Limitations}
This work focuses on building a benchmark for analyzing the robustness of TQA systems in three fine-grained aspects. In terms of the reasoning type, we mainly discuss numerical reasoning. However, other types of reasoning, e.g., commonsense reasoning \heike{or} temporal reasoning can also occur during \wei[]{the aggregation phase of} TQA. Future studies can explore these aspects and extend our benchmark. We build our benchmark with English TQA datasets, this could also be extended by incorporating TQA datasets in other languages. 
\wei[]{Additionally, though we report statistics about percentages of numeric and non-numeric values when removing relevant cells, our benchmark does not distinguish these two value types explicitly. Future work could extend our benchmark on this aspect to explore how non-numeric value changes affect model performance. }
%\blue{May not be the best way to look at impact of relevant cells on the answer. It would have beeninteresting to seen how non-numeric value changes affect model performance. Numerical values are (IMO) not verysusceptible to model internal knowledge whereas non-numeric factual information is notoriously replicated in LLMs.}\annetodo{from review, I agree, should be added here (though difficult to address)}

\section*{Ethical Considerations}
The development sets of the source datasets we use: WTQ \cite{Pasupat2015CompositionalSP}, WikiSQL \cite{Zhong2017Seq2SQLGS}, SQA \cite{Iyyer2017SearchbasedNS} and TAT \cite{Zhu2021TATQAAQ} are publicly available under the licenses of \textsc{CC-BY-SA-4.0}\footnote{https://creativecommons.org/licenses/by-sa/4.0/}, \textsc{BSD-3 Clause}\footnote{https://opensource.org/license/bsd-3-clause/}, \textsc{MIT}\footnote{https://opensource.org/license/mit/} and \textsc{MIT}, respectively. These licenses all permit us to compose, modify, publish, and distribute additional annotations upon the original dataset. Experiments in this paper are run on a single NVIDIA Tesla V100-32G GPU. Benchmark and code will be released along with the paper.
For annotation tasks where humans are involved, we recruit 5 undergraduate students (3 females and 2 males) studying Linguistics in China. All 5 annotators voluntarily participate in the annotation tasks. Three out of four of our annotation tasks requires less than 2 hours to finish. The other one took two weeks and we suggest annotators to spend less than three hours per day to ensure enough rest.

\bibliography{anthology,custom}
\newpage
\appendix
\clearpage
\section{Appendix}
\label{appendix}

\subsection{Source Dataset Overview}
\label{source_dataset_overview}
\begin{table}[!h]
\resizebox{0.9\columnwidth}{!}{
  \begin{tabular}{lllll}
    \toprule  
    \textbf{Name} & \textbf{Domain} & \textbf{Feature} & \textbf{\#Questions}&\textbf{\#Table} \\
    \midrule
    WTQ & Wikipedia & complex QA  & 2831 & 346  \\ \midrule
    WikiSQL & Wikipedia & simple QA & 8418 & 2628  \\ \midrule
    SQA &Wikipedia  & conversational QA & 2265 & 148  \\ \midrule
    TAT & Finance &  complex QA  & 1668  & 278 \\ 
    \bottomrule
  \end{tabular}%
  }
 \caption{Overview of development set of source data.}
  \label{source_dataset}
\end{table}

\subsection{Human Annotations}
\label{annotations}
Three annotators, all undergraduate students studying Linguistics in China, were involved in the annotation process. 

\paragraph{To distinguish between extraction and reasoning questions.} They are told to classify whether a question is of type extraction or reasoning after giving the definitions and examples of each question type. The instructions are listed as follow:
\begin{itemize}
    \item Thank you for participating in this annotation task!
In this task, you will be given a pair of question, answer and table and to determine if a question is extraction-based [1] or reasoning-based [0], or neither of them [-1].
We define extraction-based as: answers of the questions can be retrieved from table without numerical reasoning abilities. For instance, for the question: what is the sale number of 2018, given the rows showing the year and the column showing the sale number, the answer can be obtained by looking at certain cells without needing operations on numerical levels.
We define reasoning-based as: answers of the questions should be firstly retrieved from table and then numerical reasoning abilities are needed to obtain the answer. For instance, for the question: how many countries have received 4 goals. Entries of countries receiving 4 goals should be retrieved, then, counting is needed to obtain the answer. Common numerical operations are: counting, comparing (date or number), summing, averaging, subtracting, etc. 
If you find the question unanswerable given the table information, please put a [-1] in the annotation field.
\end{itemize}

Agreement on the binary question type classification of the 200 questions amounts to a $\kappa$-score \citep{Fleiss1971MeasuringNS} of 0.85, which shows good agreement among the annotators \cite{landis1977application}.
%\annetodo{see comment before}

\paragraph{To test the validity of relevant cells annotations.} 
We randomly sample 100 annotations for WTQ and ask the same 3 annotators to decide if removing the relevant cells prevents one from answering the questions or not. %\annetodo{is this binary test what you measure Kappa for? if so, we need to move this info to the main part of the paper}
The instruction is listed as follows:
\begin{itemize}
    \item Thank you to participate in this annotation task!
In this task, you will be given a pair of question, answer, table and relevant cells from the table to determine if removing the relevant cells prevent one from answering the question (and deriving the same correct answers) or not.
If you find removing the relevant cells prevent one from answering the question, please put [1] in the annotation field. Otherwise, please put [0]. If you find a question unanswerable or strange, please put [-1] in the annotation field. 
\end{itemize}
The Fleiss’ $\kappa$-score is 0.53, which shows moderate agreement.%\annetodo{isn't there a different number in the main part of the paper? please check}

\subsection{Eliminating Questions Related to Table Structures}
\label{eliminate_position}
We collect common positional prepositions (suggesting positions) and ordinal numbers (suggesting order) in English.
The list we use is as follows: ['first', 'second','third', 'last', 'top', 'bottom', 'before', 'previous', 'latter', 'after', 'next', 'below', 'above'].
We eliminate questions containing words in the list.

\subsection{LLMs prompt}
\label{llm_prompts}
\begin{figure}[!h]
    \centering
    \includegraphics[width=0.8\columnwidth]{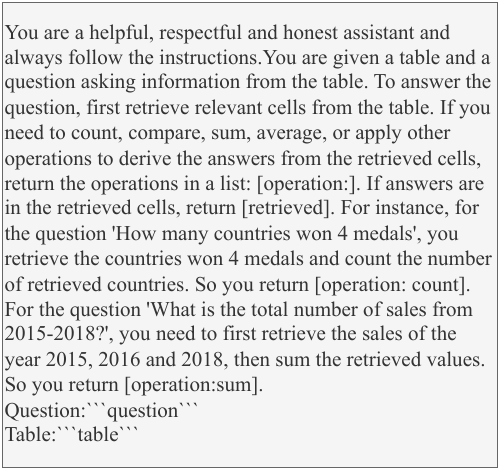}

    \caption{LLaMA2 Prompt for classifying extraction and reasoning questions.}
    \label{llm_selection_prompt}
\end{figure}

\subsection{Training/Prompting Details}
\label{train_prompt_details}
%\annetodo{I would combine this with the previous subsection, as it answers parts of the questions I had above}
Similar as in \citet{Zhao2023RobuTAS}, we fine-tuned the Large version of pre-trained end-to-end TQA systems for 20 epochs and use the random-split-1-train if multiple train-dev splits are provided. 
For TAT, the training epochs is increased to 50 epochs.
We then split the training set further into training and development sets with the ratio of 80:20. 
The batch size we use are 32 and gradient accumulation is 4.
The best model is selected based on validation loss. 
As for fine-tuning LLaMA2, we use the 7B chat-hf version. We use the code provided in LLaMA-Factory\footnote{https://github.com/hiyouga/LLaMA-Factory} and keeps the default settings of all parameters in the code. 
In terms of GPT-3.5, we use three-shots demonstrations. The demonstrations can be found in \aref{prompt_gpt_llama}.
As for Binder, since the systems used in the original paper (Codex) is no longer provided by OpenAI, we use GPT-3.5 as the backbone. 
The performance might degrade due to the switch, as what is observed in \citet{Zhang2023ReAcTableER}.
%\annetodo{mention here what that means for accuracy?} 

\lstset{%
  language=[LaTeX]TeX,
  backgroundcolor=\color{gray!25},
  basicstyle=\ttfamily,
  breaklines=true,
  columns=fullflexible
}

\subsection{Results for Robustness Against Table Structural Change for Different Datasources}
\label{appendix_tsc_results}
The following four tables show the exact match difference and variation percentage for each subset on retrieval robustness against table structure changes perturbations. 

\definecolor{babyblue}{RGB}{204,229,255}

\begin{table*}[t]
\centering
\resizebox{\textwidth}{!}{%
  \begin{tabular}{lcccccccccccc}
    \toprule
    \multirow{2}*{\textbf{Ps}} &
      \multicolumn{6}{c}{\textbf{Exact match difference}} &
      \multicolumn{6}{c}{\textbf{Variation percentage}} \\
      \cmidrule(lr){2-7}
      \cmidrule(lr){8-13}
      & {TAPEX} & {OmniTab} & {TaPas} & {GPT-3.5} & {LLaMA2} & {Binder}& {TAPEX} & {OmniTab} & {TaPas} & {GPT-3.5} & {LLaMA2} & {Binder} \\
      \cmidrule(lr){1-7}
      \cmidrule(lr){8-13}
    NP & 84 & \cellcolor{babyblue} 84.57 & 69.71 & 77.71 & 63.43 & 72.57 & 0 & 0 & 0 & 0 & 0 & 0 \\
    \midrule
    R-SA & $-3.09\pm0.46$ & $-1.03\pm0.98$ & $-0.91\pm1.06$ & \cellcolor{babyblue} $-0.57\pm1.20$ & $-3.43\pm2.20$ & $-2.86\pm1.30$ &  $7.43\pm1.14$  & \cellcolor{babyblue} $4.46\pm0.98$ & $7.54\pm1.11$ & $11.31\pm1.42$ & $16.46\pm2.03$ &  $9.03\pm1.51$ \\
    %R-TR & $-1.71\pm1.30$ & \cellcolor{babyblue} 0.46\pm0.43 & $-1.14\pm1.70$ & $-0.91\pm1.18$ &$-0.69\pm1.82$  &$-3.66\pm1.33$ & \cellcolor{babyblue} 4.00\pm0.96 & $4.11\pm0.43$  & $6.17\pm1.17$ & $10.06\pm1.79$ & $12.34\pm1.56$ & $10.74\pm2.62$ \\
    R-TT & $-2.17\pm1.17$ & \cellcolor{babyblue} $0.00\pm0.36$ &$ 0.57\pm0.89$ & $-0.69\pm1.71$ & $1.26\pm1.27$& $-3.89\pm1.75$ & $3.31\pm0.23$ & \cellcolor{babyblue} $2.51\pm0.58 $ & $4.46\pm0.23$ & $12.80\pm1.83$ &$10.40\pm0.91$  & $11.43\pm2.96$ \\
    R-TM & $-1.03\pm0.56$ & \cellcolor{babyblue} $0.00\pm0.81$ & $-1.71\pm0.36$ & $-1.49\pm0.78$ & $-2.40\pm0.56$ & $-3.54\pm2.21$ & \cellcolor{babyblue} $1.71\pm0.00$ & $3.43\pm0.36$ & $3.54\pm1.11$ & $11.09\pm1.78$ & $9.71\pm0.72$ & $10.86\pm2.91$\\
    R-TB & $-2.63\pm0.69$ & $-1.03\pm0.84$ & $-0.69\pm0.43$ & \cellcolor{babyblue} $-0.46\pm1.37$ & $-2.97\pm1.75$ & $-3.43\pm1.20$ & $4.23\pm1.43$ & \cellcolor{babyblue} $3.09\pm0.28$ & $4.57\pm0.36$ & $10.51\pm0.86$ & $14.40\pm0.43$ & $10.06\pm2.19$\\
          \midrule
    C-SA & $-4.91\pm0.93$& \cellcolor{babyblue} $-0.91\pm1.28$ & $-2.51\pm1.18$ & $-7.31\pm1.51$ & $-6.74\pm2.92$ & $-4.11\pm1.75$ & $11.09\pm2.30$ & \cellcolor{babyblue} $5.94\pm0.78$ & $6.17\pm0.84$ & $18.51\pm1.93$& $11.09\pm2.30$ &$10.97\pm1.42$ \\
    %C-TR & $-5.26\pm0.98$ & \cellcolor{babyblue} -0.11\pm1.67 & $-1.94\pm1.00$ & $-5.14\pm1.25$ & $-2.17\pm2.94$ & $-3.43\pm1.14$ & $7.77\pm2.57$& \cellcolor{babyblue} 5.60\pm1.32 & \cellcolor{babyblue} 5.60\pm1.27 & $16.57\pm1.98$ & $13.83\pm2.24$ & $10.29\pm2.04$\\
    C-TF& $-3.77\pm0.46$ & \cellcolor{babyblue} $-1.03\pm0.56$ & $-1.37\pm0.28$ & $-5.60\pm2.21$ & $-5.46\pm3.85$  & $-2.86\pm1.20$  & $6.51\pm1.00$ & \cellcolor{babyblue} $3.77\pm1.12$ & $4.34\pm0.28$ & $16.11\pm2.79$ & $15.77\pm2.57$ & $10.86\pm2.48$ \\
    C-TB & $-6.97\pm2.71$ & \cellcolor{babyblue} $-1.26\pm0.56$ & $-2.40\pm0.91$ & $-6.17\pm1.11$ & $-4.80\pm1.47$ & $-2.97\pm2.41$ & $9.49\pm2.44$ & $5.83\pm0.56$ & \cellcolor{babyblue} $4.46\pm1.37$ & $16.23\pm2.00$ & $12.91\pm2.00$ & $10.97\pm1.93$\\ 
      \midrule
     Tr & -65.14 & -64.57 & -54.86 & \cellcolor{babyblue} -18.86 & -34.29 & -70.86  & 68.57 & 70.29 & 61.71 & \cellcolor{babyblue} 25.71 & 42.29 & 70.86\\
    \bottomrule
  \end{tabular}}
\caption{WTQ: system performance on extraction data set. Ps stands for perturbations. NP stands for no perturbation. R, C and Tr stand for row, column and transpose perturbations. SA, TT, TF, TM and TB stand for shuffle all, target top, target front, target middle, target bottom/back respectively. We shade the best performances (minimal absolute values) with blue.}
\label{wtq_factoid}
\end{table*}

\begin{table*}
\centering
\resizebox{\textwidth}{!}{%
  \begin{tabular}{lcccccccccccc}
    \toprule
        \multirow{2}*{\textbf{Ps}} &
      \multicolumn{6}{c}{\textbf{Exact match difference}} &
      \multicolumn{6}{c}{\textbf{Variation percentage}}  \\
      \cmidrule(lr){2-7}
      \cmidrule(lr){8-13}
      & {TAPEX} & {OmniTab} & {TaPas} & {GPT-3.5} & {LLaMA2} & {Binder}& {TAPEX} & {OmniTab} & {TaPas} & {GPT-3.5} & {LLaMA2} & {Binder} \\
      \cmidrule(lr){1-7}
      \cmidrule(lr){8-13}
    NP & 89.27 & 91.33 & \cellcolor{babyblue} 92.59 & 63.06 & 56.45 &66.70  & 0 & 0 & 0 & 0 & 0 & 0 \\
    \midrule
    R-SA & $-0.08\pm0.04$ &\cellcolor{babyblue} $0\pm0.06$ & $0.08\pm0.04$ & $-0.27\pm0.13$ & $-1.82\pm0.22$ & $-1.14\pm0.11$ & $1.08\pm0.06$ &\cellcolor{babyblue} $0.63\pm0.02$ & $0.63\pm0.04$&$10.37\pm1.11$ & $12.26\pm0.88$ & $9.21\pm0.63$ \\
    %Target random & $-0.08\pm0.06$ & $-0.08\pm0.12$ & $-0.01\pm0.03$ & $-0.11\pm0.14$ & $-1.35\pm0.66$ & $-1.17\pm0.11$ & $0.84\pm0.12$ & $0.37\pm0.02$  & $0.44\pm0.03$ & $11.89\pm0.89$& $12.37\pm1.66$ &$11.28\pm0.91$  \\
    R-TT & \cellcolor{babyblue}$0.02\pm0.01$ & $0.04\pm0.02$ & $0.03\pm0.03$ & $0.12\pm0.07$ & $1.01\pm0.52$ & $-0.93\pm0.05$ & $0.57\pm0.06$ & \cellcolor{babyblue}$0.30\pm0.01$ & $0.42\pm0.02$ & $11.50\pm0.56$ & $13.75\pm1.21$ & $10.77\pm0.94$  \\
    R-TM & $-0.08\pm0.04$ & $0.11\pm0.01$ & \cellcolor{babyblue}$0.04\pm0.02$ & $-1.04\pm0.11$ & $-1.24\pm0.44$ &$-1.23\pm0.05 $& $0.57\pm0.06$ & \cellcolor{babyblue}$0.30\pm0.02$ & $0.44\pm0.02$ & $13.61\pm0.11$ & $14.4\pm0.42$ &$12.84\pm0.47$  \\
    R-TB & $-0.12\pm0.03$ & \cellcolor{babyblue}$-0.06\pm0.04$ & $-0.08\pm0.03$ &$-0.22\pm0.13$&$-1.88\pm0.79$ &$-1.15\pm0.07$&$0.68\pm0.05$ & \cellcolor{babyblue}$0.43\pm0.02$ & $0.51\pm0.08$ & $12.24\pm0.77$& $14.61\pm0.68$ & $10.45\pm0.79$\\
    \midrule
     C-SA & $-0.45\pm0.12$&$ -0.12\pm0.04$& \cellcolor{babyblue}$-0.06\pm0.06$ & $-2.22\pm0.28$ & $-1.53\pm0.26$& $-2.06\pm0.22$ & $2.35\pm0.10$ & $1.73\pm0.12$ & \cellcolor{babyblue}$1.02\pm0.02$ & $19.09\pm0.59$ & $15.43\pm0.90$ & $16.72\pm0.85$\\
    %Target random & $-0.47\pm0.24$ & $-0.05\pm0.05$ & $-0.03\pm0.03$& $-2.26\pm0.22$ & $-1.93\pm0.41$ &$-2.17\pm0.18$  & $1.55\pm0.10$& $0.79\pm0.05$ & $0.62\pm0.07$ & $17.68\pm0.72$ & $16.91\pm0.78$ & $14.59\pm0.88$\\
    C-TF & $-0.60\pm0.03$ & $0.06\pm0.01$ & \cellcolor{babyblue}$-0.01\pm0.01$ & $-3.26\pm0.37$ & $-4.01\pm0.52$ &$-3.03\pm0.22$  & $1.55\pm0.14$ & $0.79\pm0.05$ & \cellcolor{babyblue}$0.62\pm0.07$ & $18.26\pm0.60$ & $15.40\pm0.37$ & $13.11\pm0.79$ \\
    C-TB & $-0.26\pm0.10$ & $-0.20\pm0.07$ & \cellcolor{babyblue}$-0.11\pm0.04$ & $-3.43\pm0.33$ & -$2.37\pm0.77$& $-3.02\pm0.21$ & $1.17\pm0.05$ & $0.57\pm0.09$ & \cellcolor{babyblue}$0.43\pm0.06$ & $17.14\pm0.52$ & $15.26\pm0.58$ & $11.47\pm0.61$\\ 
      \midrule
     Tr & -76.13& -78.72 & -88.13 & \cellcolor{babyblue}-15.95 & -32.72 & -56.07  & 76.99 & 79.34 & 88.17 & \cellcolor{babyblue}30.15 & 40.72 &73.69 \\
    \bottomrule
  \end{tabular}
}
\caption{WikiSQL: system performance on extraction data set. Ps stands for perturbations. NP stands for no perturbation. R, C and Tr stand for row, column and transpose perturbations. SA, TT, TF, TM and TB stand for shuffle all, target top, target front, target middle, target bottom/back respectively. We shade the best performances (minimal absolute values) with blue.}
\label{wikisql_factoid}
\end{table*}

\begin{table*}[!h]
\centering
\resizebox{\textwidth}{!}{%
  \begin{tabular}{lcccccccccccc}
    \toprule
    \multirow{2}*{\textbf{Ps}} & \multicolumn{6}{c}{\textbf{Exact match difference}} & \multicolumn{6}{c}{\textbf{Variation percentage}}  \\
      \cmidrule(lr){2-7}
      \cmidrule(lr){8-13}
      & {TAPEX} & {OmniTab} & {TaPas} & {GPT-3.5} & {LLaMA2} & {Binder}& {TAPEX} & {OmniTab} & {TaPas} & {GPT-3.5} & {LLaMA2} & {Binder} \\
      \cmidrule(lr){1-7} \cmidrule(lr){8-13}
    NP & 67.39 & 70.29 & 68.84 & \cellcolor{babyblue} 71.74 & 57.25 & 63.89 & 0 & 0 & 0 & 0 & 0 & 0 \\
    \midrule
    R-SA & \cellcolor{babyblue}$-1.21\pm1.46$ & $-1.45\pm0.57$ & $-2.90\pm1.18$ & $-1.45\pm0.78$ & $-2.17\pm0.98$ & $-1.39\pm0.90$ &  $5.56\pm1.18$  & \cellcolor{babyblue}$4.35\pm1.17$ & $7.25\pm1.69$ & $8.78\pm1.40$ & $9.98\pm1.04$ &$8.54\pm1.14$  \\
    %Target random & $-1.48\pm1.09$ & $-1.38\pm0.23$ & $-3.35\pm0.59$ & $-2.05\pm1.23$ & $-1.58\pm0.89$ &  $-2.03\pm0.52$ & $5.31\pm1.02$ & $4.83\pm0.23$  & $7.73\pm0.79$ & $10.14\pm1.23$& $10.07\pm1.57$ & $10.43\pm1.77$ \\
    R-TT & $0.45\pm1.12$ & \cellcolor{babyblue}$0.24\pm0.90$ & $-1.45\pm0.59$ &$-0.72\pm0.97$ & $-2.17\pm1.24$ & $-1.43\pm0.57$ & $3.86\pm1.37$& \cellcolor{babyblue}$2.66\pm0.34$ & $3.38\pm0.34$ & $10.82\pm1.87$ & $9.42\pm1.24$ & $9.43\pm1.83$ \\
    R-TM & $-3.48\pm0.49$& \cellcolor{babyblue}$-3.42\pm0.34$ & $-5.07\pm0.59$ & $-5.12\pm0.88$ & $-3.71\pm1.53$ & $-4.84\pm1.01$ & $3.86\pm0.49$ & \cellcolor{babyblue}$3.86\pm0.34$ & $5.56\pm0.34$ & $9.37\pm1.22$ & $10.22\pm0.93$ & $9.46\pm1.27$ \\
    R-TB & $-4.66\pm0.73$ & \cellcolor{babyblue}$-4.59\pm0.34$& $-4.83\pm0.34$ & $-5.80\pm0.43$ &$-4.90\pm1.57$ & $-6.12\pm0.67$ & \cellcolor{babyblue}$3.42\pm1.70$& $4.34\pm0.97$& $6.02\pm0.35$ & $11.39\pm1.66$ & $13.68\pm2.54$ & $10.87\pm1.21$\\
          \midrule
     C-SA & $-3.14\pm1.26$& \cellcolor{babyblue}$-0.97\pm1.07$ & $-3.14\pm0.86$ & $-5.21\pm1.34$& $-7.65\pm1.70$ & $-3.79\pm1.05$& $12.32\pm1.57$ & $8.21\pm1.37$ &$7.97\pm1.05$ & $14.21\pm1.87$& $16.21\pm1.63$ &\cellcolor{babyblue}$7.69\pm1.37$ \\
    %Target random & $-4.35\pm0.77$ & $-3.14\pm0.68$ & $-1.45\pm0.59$ & $-5.34\pm1.66$ & $-3.91\pm0.80$ & $-4.17\pm1.70$ & $11.71\pm0.97$& $7.74\pm0.17$ & $7.59\pm0.62$ & $12.87\pm1.93$ & $10.29\pm2.07$ &$8.57\pm1.28$ \\
    C-TF &$-2.17\pm1.02$ & $-2.17\pm0.96$ &\cellcolor{babyblue}$-0.24\pm0.31$ & $-6.38\pm1.54$ & $-7.57\pm1.39$ & $-5.25\pm0.77$ & $7.00\pm1.34$ & \cellcolor{babyblue}$5.17\pm1.02$ &$5.24\pm1.71$ & $14.49\pm1.92$ &$13.14\pm2.38$ &$6.29\pm1.51$  \\
    C-TB & $-2.90\pm1.57$ & $-3.14\pm0.68$ & \cellcolor{babyblue}$-1.21\pm0.68$ & $-5.11\pm1.79$ & $-4.38\pm1.13$& $-3.58\pm1.11$ &$10.14\pm1.05$ & $7.49\pm0.68$ &\cellcolor{babyblue}$3.66\pm0.68$ & $11.62\pm2.78$ & $12.80\pm2.25$&$5.57\pm1.12$ \\ 
      \midrule
     Tr & -51.45& -58.70 & -63.04 & \cellcolor{babyblue}-21.01 & -34.78 & -53.48  & 57.25 & 60.14 & 64.49 & \cellcolor{babyblue}25.36 & 39.13&68.24 \\
    \bottomrule
  \end{tabular}}
\caption{SQA: system performance on extraction data set. Ps stands for perturbations. NP stands for no perturbation. R, C and Tr stand for row, column and transpose perturbations. SA, TT, TF, TM and TB stand for shuffle all, target top, target front, target middle, target bottom/back respectively. We shade the best performances (minimal absolute values) with blue.}
\label{sqa_factoid}
\end{table*}

\begin{table*}[!h]
\centering
\resizebox{\textwidth}{!}{
  \begin{tabular}{lcccccccccccc}
    \toprule
     \multirow{2}*{\textbf{Ps}} &\multicolumn{6}{c}{\textbf{Exact match difference}} &\multicolumn{6}{c}{\textbf{Variation percentage}} \\
      \cmidrule(lr){2-7} \cmidrule(lr){8-13}
      & {TAPEX} & {OmniTab} & {TaPas} & {GPT-3.5} & {LLaMA2} & {Binder}& {TAPEX} & {OmniTab} & {TaPas} & {GPT-3.5} & {LLaMA2} & {Binder} \\
      \cmidrule(lr){1-7} \cmidrule(lr){8-13}
    NP &78.65 & \cellcolor{babyblue}82.75 &56.14 & 74.56 &67.25 & 67.02 & 0 & 0 & 0 & 0 & 0 & 0 \\
    \midrule
    R-SA & $-3.43\pm0.41$ & $-2.71\pm2.07$ & $-4.02\pm0.72$ & $-2.06\pm1.09$ & $-3.05\pm1.09$ &\cellcolor{babyblue} $-2.05\pm1.41$ &  $10.53\pm1.89$  & \cellcolor{babyblue}$8.77\pm1.24$ & $10.04\pm1.89$ & $15.50\pm1.80$ & $15.56\pm1.80$ &$16.73\pm1.09$  \\
    %Target & $-2.05\pm0.83$ & $-2.63\pm0.72$ & $-1.75\pm1.24$ & $-5.19\pm2.52$ & $-5.58\pm1.83$ &  $-4.03\pm1.72$ & $9.36\pm2.19$ & $6.73\pm1.24$  & $6.14\pm0.41$ & $14.62\pm2.41$& $13.36\pm1.83$ & $14.39\pm1.72$ \\
    R-TT & \cellcolor{babyblue} $0.29\pm0.41$ & $0.29\pm0.83$ & $-4.09\pm0.83$ &$-5.85\pm2.19$ & $-3.05\pm2.07$ & $-4.46\pm2.41$ & $2.63\pm1.24$&\cellcolor{babyblue} $2.05\pm1.65$ & $4.09\pm0.83$ & $10.53\pm3.79$ & $10.97\pm2.89$ & $10.97\pm1.49$ \\
    R-TM & $-3.39\pm0.82$& \cellcolor{babyblue}$-2.72\pm0.41$ & $-2.88\pm1.24$ & $-3.74\pm1.65$ & $-2.85\pm0.41$ & $-2.88\pm0.72$ & $5.56\pm1.65$ & \cellcolor{babyblue}$4.09\pm1.01$ & $6.14\pm1.24$ & $9.01\pm2.56$ & $10.39\pm1.89$ & $8.56\pm1.09$ \\
    R-TB & $-3.06\pm2.73$ & \cellcolor{babyblue}$-2.59\pm0.34$& $-3.83\pm0.34$ & $-3.80\pm1.43$ &$-3.90\pm2.57$ & $-3.12\pm1.67$ & $6.42\pm2.70$& \cellcolor{babyblue}$4.27\pm1.97$& $7.13\pm2.35$ & $11.19\pm2.37$ & $13.68\pm2.54$ & $9.95\pm2.20$\\
    \midrule
     C-SA & $-4.97\pm2.07$&\cellcolor{babyblue} $-4.92\pm2.89$ & $-4.97\pm2.41$ & $-5.70\pm2.41$& $-5.58\pm2.47$ & $-5.29\pm1.49$& \cellcolor{babyblue}$6.14\pm1.24$ & $6.43\pm0.41$ &$7.89\pm1.24$ &$11.71\pm3.80$ & $12.60\pm2.07$& $10.56\pm2.54$ \\
    %Target random & $-3.63\pm1.24$ & $-2.46\pm2.07$ & $-2.46\pm1.65$ & $-4.74\pm2.41$ & $-4.14\pm2.41$ & $-4.52\pm2.72$ & $8.71\pm2.97$& $7.89\pm1.24$ & $6.73\pm0.41$ & $13.80\pm1.65$ & $11.77\pm2.24$ &$10.68\pm2.14$ \\
    C-TF &$-5.31\pm1.24$ & $-4.68\pm1.65$ &\cellcolor{babyblue}$-4.31\pm0.41$ & $-4.91\pm1.24$ & $-4.80\pm1.65$ & $-4.75\pm0.79$ & $6.73\pm2.07$ & $5.11\pm1.65$ &\cellcolor{babyblue}$4.82\pm1.41$& $13.93\pm2.24$ &$12.89\pm2.80$ &$10.85\pm0.83$  \\
    C-TB & $-4.97\pm2.07$ & $-4.88\pm1.03$ & \cellcolor{babyblue}$-4.75\pm0.82$ & $-5.50\pm1.09$ & $-5.58\pm2.41$& $-5.75\pm1.24$ &$7.31\pm1.41$ & $7.27\pm0.81$ &\cellcolor{babyblue}$6.43\pm0.83$ & $14.93\pm3.72$ & $12.34\pm2.70$&$12.85\pm2.41$ \\ 
    \midrule
     Tr & -28.71& -29.36 & -37.89 & \cellcolor{babyblue}-13.45 & -20.51 & -27.02  & 35.73 & 31.05 & 47.98 & \cellcolor{babyblue}23.39 & 31.53&38.02 \\
    \bottomrule
  \end{tabular}}
\caption{TAT: system performance on extraction data set. Ps stands for perturbations. NP stands for no perturbation. R, C and Tr stand for row, column and transpose perturbations. SA, TT, TF, TM and TB stand for shuffle all, target top, target front, target middle, target bottom/back respectively. We shade the best performances (minimal absolute values) with blue.}
\label{tat_factoid}
\end{table*}

\subsection{LLM prompts for GPT3.5}
\begin{figure*}[!ht]
    \centering
    \includegraphics[width=1.\textwidth]{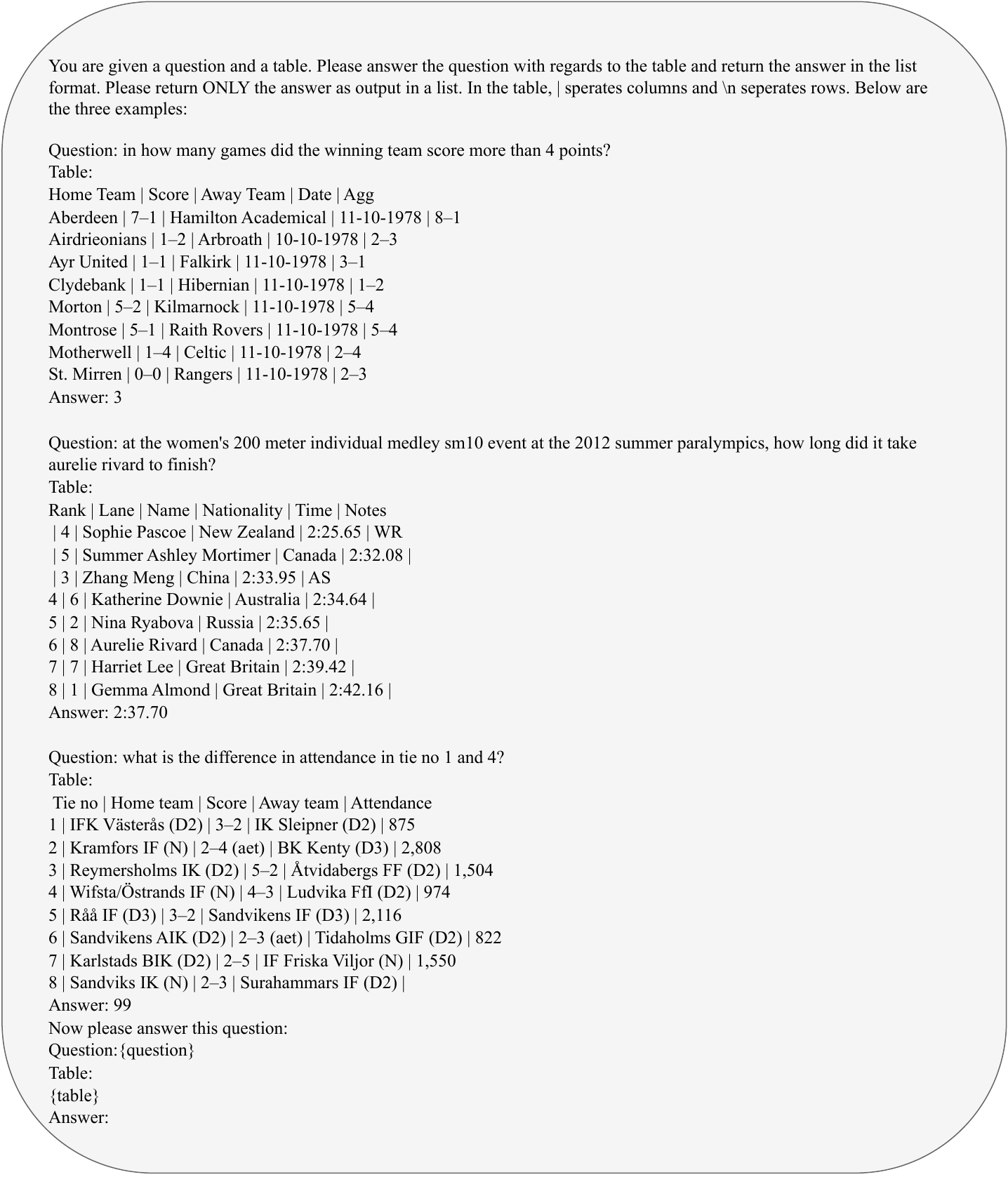}

    \caption{Prompt for GPT-3.5 for WTQ.}
    \label{llm_prompt_experiment}
\end{figure*}

\begin{figure*}[!ht]
    \centering
    \includegraphics[width=1.\textwidth]{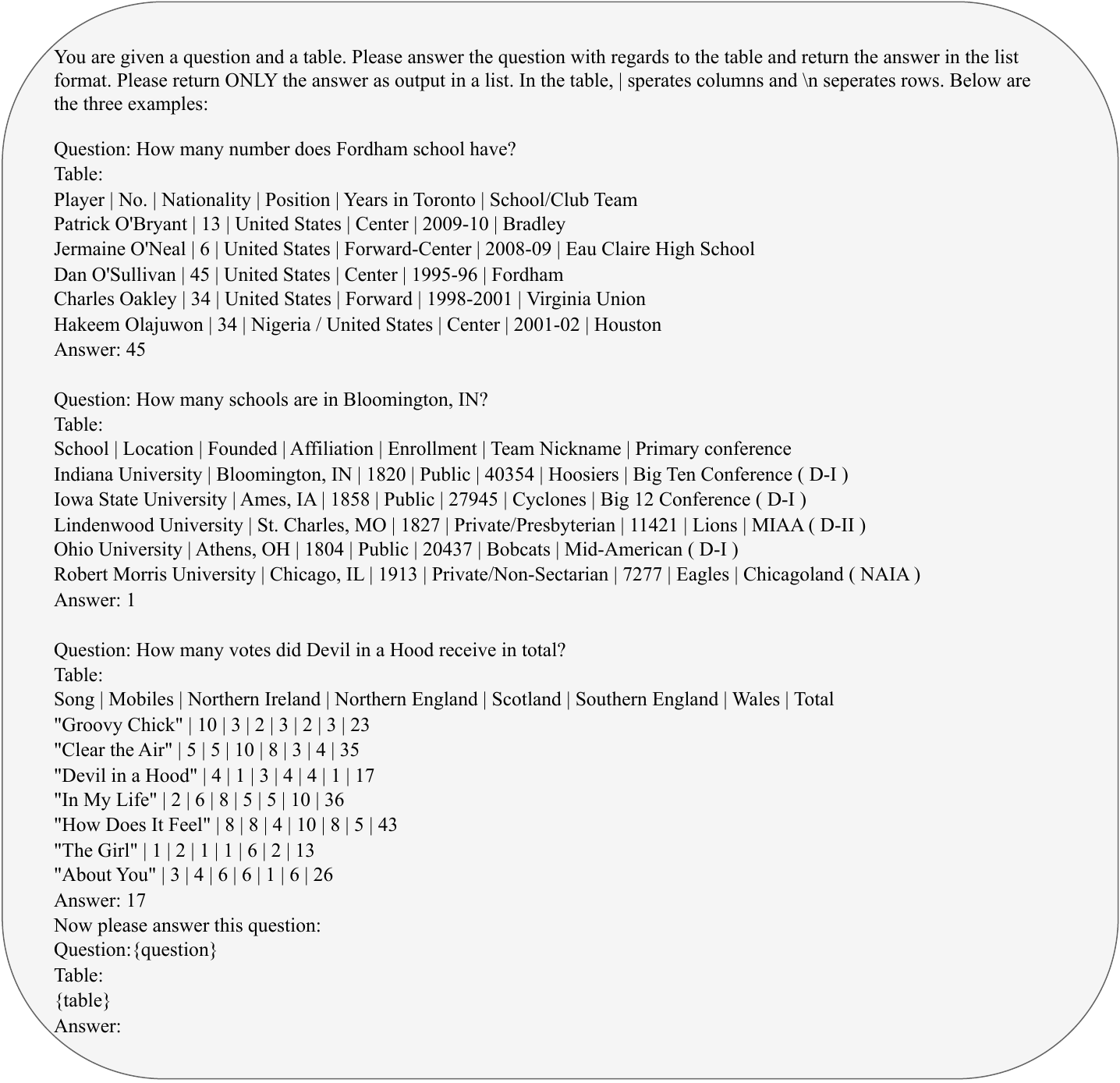}

    \caption{Prompt for GPT-3.5 for WikiSQL.}
    \label{llm_prompt_experiment_wikisql}
\end{figure*}

\begin{figure*}[!ht]
    \centering
    \includegraphics[width=1.\textwidth]{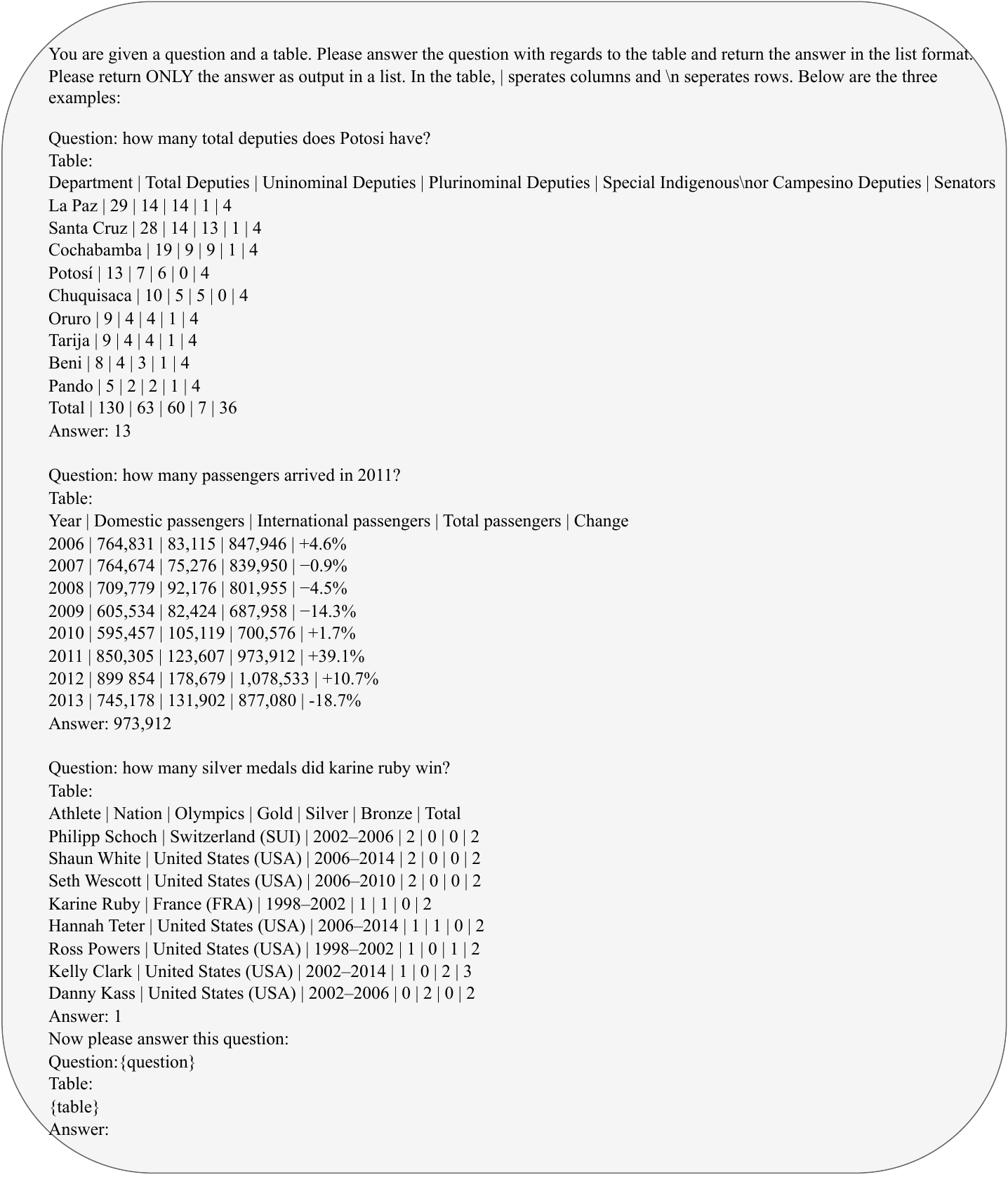}

    \caption{Prompt for GPT-3.5 for SQA.}
    \label{llm_prompt_experiment_sqa}
\end{figure*}

\begin{figure*}[!ht]
    \centering
    \includegraphics[width=1.\textwidth]{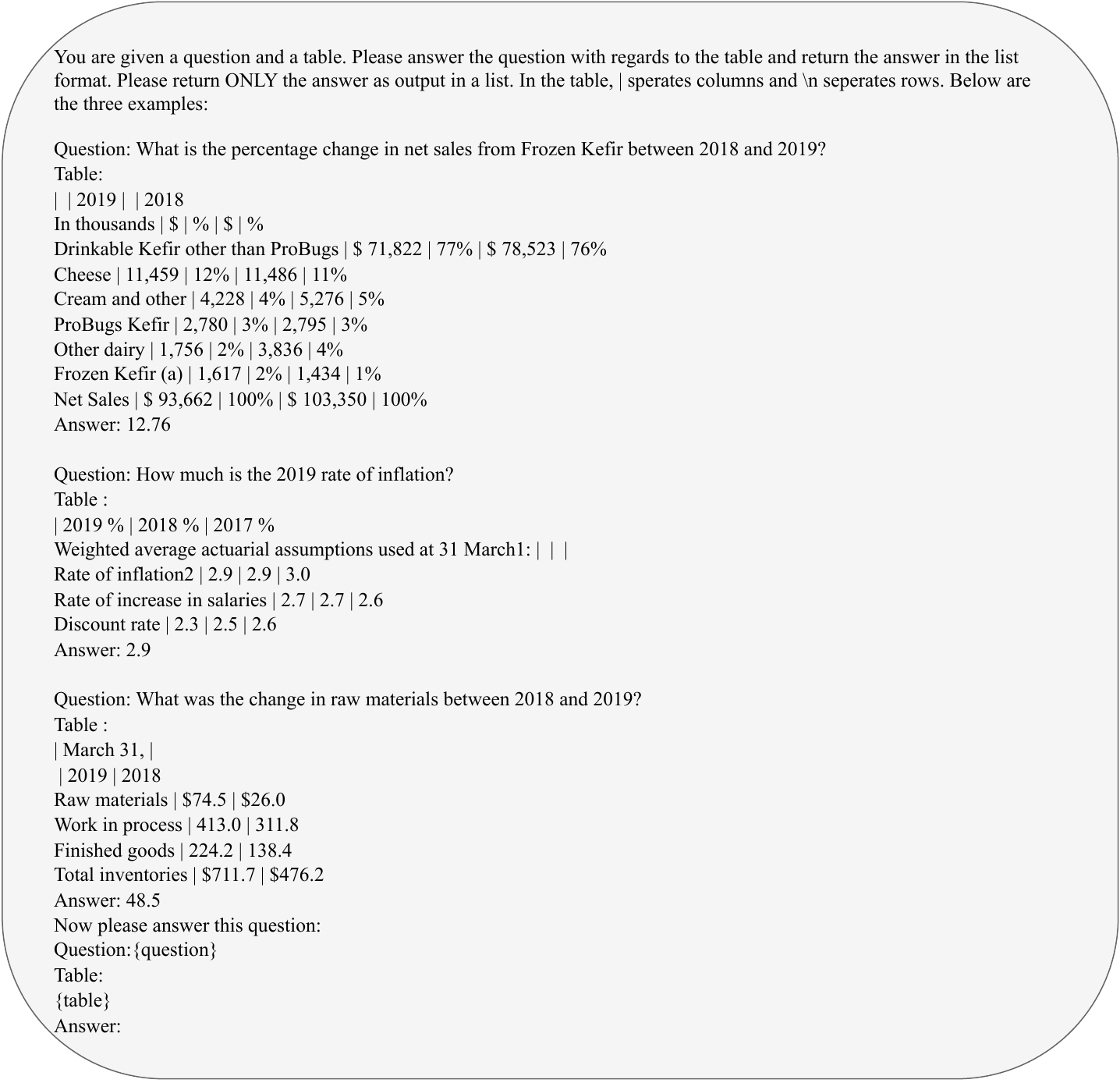}

    \caption{Prompt for GPT-3.5 for TAT.}
    \label{llm_prompt_experiment_tat}
\end{figure*}

Figure \ref{llm_prompt_experiment}, Figure \ref{llm_prompt_experiment_wikisql}, Figure \ref{llm_prompt_experiment_sqa} and Figure \ref{llm_prompt_experiment_tat} shows the prompts we used for prompting GPT-3.5 for WTQ, WikiSQL, SQA and TAT, respetively.
\label{prompt_gpt_llama}

\end{document}